\documentclass[]{fairmeta}

\usepackage{colortbl}
\usepackage{xcolor}
\usepackage{enumitem}
\definecolor{darkgreen}{rgb}{0,0.5,0}
\definecolor{light-gray}{gray}{0.9}
\definecolor{comm}{gray}{0.5}
\definecolor{denim}{rgb}{0.08, 0.38, 0.74}
\definecolor{mint}{RGB}{220,245,230}
\definecolor{pistachio}{RGB}{236,246,228} 
\usepackage[table,svgnames]{xcolor}
\usepackage{hyperref}
\usepackage{amsfonts}
\hypersetup{
  colorlinks   = true,
  urlcolor     = denim,
  linkcolor    = denim,
  citecolor    = denim,
}
\usepackage[most]{tcolorbox}
\usepackage{enumitem}
\usepackage{booktabs} 
\usepackage{tabularx}
\usepackage{multirow}

\usepackage[utf8]{inputenc}
\DeclareUnicodeCharacter{2264}{\ensuremath{\le}}
\DeclareUnicodeCharacter{2265}{\ensuremath{\ge}}
\DeclareUnicodeCharacter{2013}{--}
\DeclareUnicodeCharacter{2014}{---}

\usepackage{adjustbox}
\usepackage{fvextra} % improved Verbatim with line breaking

\usepackage[table]{xcolor} % Add this to your preamble
\usepackage{subcaption}
\usepackage{graphicx}
\usepackage{cleveref}
\crefname{section}{Section}{Sections}
\crefname{subsection}{Section}{Sections}

\usepackage{inconsolata}
\usepackage{tcolorbox}
\tcbuselibrary{most}

\newtcolorbox{prompt}[1]{
    enhanced,
    left=4mm,
    right=4mm,
    top=2mm,
    bottom=2mm,
    boxsep=0mm,
    rounded corners,
    title=#1,
    fontupper=\footnotesize\linespread{0.9}\fontfamily{lmr}\selectfont,
}

\title{Autodata: An agentic data scientist to create high quality synthetic data}

\abstract{
We introduce Autodata, a general method that enables AI agents to act as data scientists who  build high quality training and evaluation data. 
We show how to train (meta-optimize) such a data scientist agent, so that it learns to create even stronger data.
We describe the overall formulation, and a specific practical implementation, Agentic Self-Instruct. We conduct experiments on computer science research tasks, legal reasoning tasks and reasoning with mathematical objects, where we obtain  improved results compared to classical synthetic dataset creation methods. 
Further, meta-optimizing the data scientist agent itself delivers an even larger performance uplift.
Agentic data creation provides a way to convert increased inference compute into higher quality model training. Overall, we believe this direction has the potential to change the way we build AI data.
}

\author{Ilia Kulikov$^\dagger$, Chenxi Whitehouse$^\dagger$, Tianhao Wu$^\dagger$, Yixin Nie$^\dagger$, Swarnadeep Saha, Eryk Helenowski, Weizhe Yuan, Olga Golovneva, Jack Lanchantin, Yoram Bachrach, Jakob Foerster, Xian Li, Han Fang, Sainbayar Sukhbaatar, Jason Weston}

\if 0
\author{Ilia Kulikov$^\dagger$}
\author{Chenxi Whitehouse$^\dagger$}
\author{Tianhao Wu$^\dagger$}
\author{Yixin Nie$^\dagger$}
\author{Swarnadeep Saha}
\author{Eryk Helenowski}
\author{Weizhe Yuan}
\author{Olga Golovneva}
\author{Jack Lanchantin}
\author{Yoram Bachrach}
\author{Jakob Foerster}
\author{Xian Li}
\author{Han Fang}
\author{Sainbayar Sukhbaatar}
\author{Jason Weston}
\fi 
\affiliation[]{FAIR at Meta~~~~~$^\dagger$\small{Joint first author}}

\date{\today}
\correspondence{\email{kulikov@meta.com}, \email{jase@meta.com}}

\begin{document}

\maketitle

\begin{figure}[h] %tbp!]
  \centering
  \includegraphics[width=0.8\linewidth]{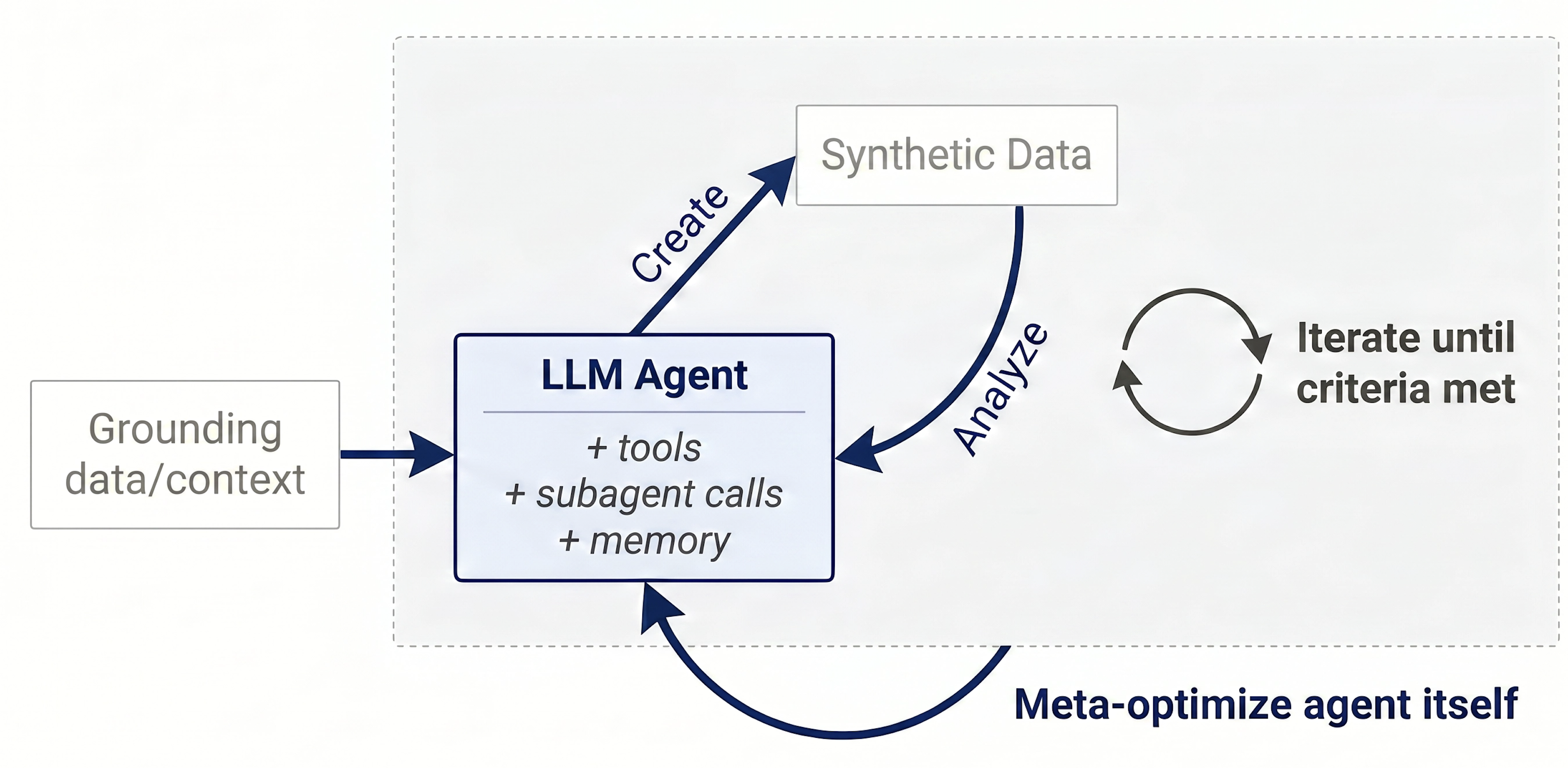}
  \caption{{\bf Autodata pipeline.} 
  The framework employs an autonomous agent that emulates the role of a data scientist, iteratively generating data, conducting qualitative inspection and quantitative performance evaluation, synthesizing insights, and updating the data-generation recipe. The agent itself can be trained to be better at the data scientist role using the same criteria used in the inner loop. This cyclical process aims to progressively enhance data quality; the diagram depicts the general workflow underlying possible instantiations.}
  \label{fig:fig1}
\end{figure}

\section{Introduction}

Progress at the AI frontier increasingly depends on high-quality training data and benchmarks that continue to challenge models. The initial foundation for training current AI systems is human-written training data. However, increasingly performance improvements are derived from synthetic data created by the model itself. Synthetic data addresses several practical challenges: it facilitates the generation of edge cases and long-tail scenarios that are underrepresented in real corpora, reduces the difficulty and latency associated with manual labeling, and can potentially produce more challenging data than the human-generated data distribution.

With the introduction of large language models (LLMs) and the ability to use in-context learning and instruction following,  Self-Instruct \citep{wang2023self} emerged as a method to create synthetic data through zero or few-shot prompting. Grounded Self-Instruct \citep{lupidi2024source2synth,yuan2025naturalreasoning} extended that to ground on documents and other sources to reduce hallucination and increase diversity. Further, methods like CoT Self-Instruct \citep{yu2025cot} extended that to use Chain-of-Thought reasoning during the generation process to help construct more complex tasks more accurately. Finally, so called ``self-challenging'' methods \citep{zhou2025self} allowed a challenger agent to interact with tools before proposing a task and accompanying evaluation functions.
However, none of these methods  control the difficulty and quality of the data directly, motivating approaches such as filtering \citep{yu2025cot}, evolution \citep{xu2024wizardlm} and refinement \citep{shah2024ai}.

In this work, we introduce Autodata, which generalizes all the above described methods. An agent acting as a data scientist is tasked with the act of constructing and curating data, performing the actions a human data scientist would take in order to create high quality data: where both building benchmark data and training data are use cases. This process includes both an initial iteration of data creation, followed by an analysis phase “eyeballing” the data as well as measuring its performance, constructing learnings, and then iterating with an improved recipe to create better data. Further, we show how to train (meta-optimize) this agentic system (outer loop) to be optimal as a data scientist (inner loop). 
While much of the recent work on autoresearch \citep{karpathy_autoresearch_2026} has concentrated on agentic methods for architectural or training recipe improvements, 
we posit that focusing on {\em data} is likely to play an equally important, if not more important, role in future progress.
 
In our experiments, we focus on a particular implementation of Autodata, Agentic Self-Instruct, and show that it provides strong results across a diverse set of tasks.
We conduct experiments on computer science research tasks, legal reasoning tasks and reasoning with mathematical objects, where in each case we obtain
improved results compared to classical synthetic dataset creation methods. Further, meta-optimizing the data scientist agent itself delivers an even larger performance uplift. 

As state-of-the-art LLMs become ever-stronger, there is concern that existing tasks or synthetic data methods cannot produce tasks that are challenging enough to make further progress. Autodata, via agentic data creation, provides a way to convert increased inference compute into higher quality model training to produce such challenging data. Hence, we believe that this direction has the potential to change the way we create new tasks and benchmarks to advance the frontier.

\section{Autodata}
\label{sec:autodata}

The  high-level design of Autodata is shown in \autoref{fig:fig1}, where various instantiations can be built from this template.
The overall loop consists of the following components.

\noindent \textbf{Data Creation.} The autodata agent grounds on some provided data (e.g. specific documents from math, legal, coding etc. or another useful data source, depending on the task) to help create the data. The agent can then use tools or existing skills/learnings it has previously acquired and inference time compute to create training or evaluation data for model training and benchmarking. Importantly, this creation step can be repeated after subsequent analysis and learnings to improve the data even further. 

\noindent \textbf{Data Analysis.} Given the data that the agent has created, it can then analyze the data for learnings on what it did right versus wrong, and how it can be improved. This could be at the level of a specific example (checking if an example is correct? high quality? challenging enough?), or potentially at the dataset level (are the samples diverse? do they improve a model if used as training data?). These learnings are fed back into the data creation process to improve the data in the next iteration, until a stopping criteria is met.

\noindent \textbf{Overall Data Scientist Loop.} The agent loops over the data creation and data analysis stages until it is satisfied with the quality of the data, and then generates a final training dataset or benchmark. This can include specific guardrails in the outer loop to prevent hacking. The agentic loop allows the model to build on top of its own learnings during these steps.

\noindent \textbf{Meta-Optimization of the Data Scientist.} The agent itself can also be optimized to be {\em better at being a data scientist}.
One way to do this is to optimize the agent harness using autoresearch \citep{karpathy_autoresearch_2026} or meta-harness \citep{lee2026meta} style optimization using the same inner loop criteria (creating better data) to guide the optimization of the outer loop (the agent optimization itself). This is depicted in the outer box of \autoref{fig:fig1}.

\begin{figure}[tbp!]
  \centering
  \includegraphics[width=0.8\linewidth]{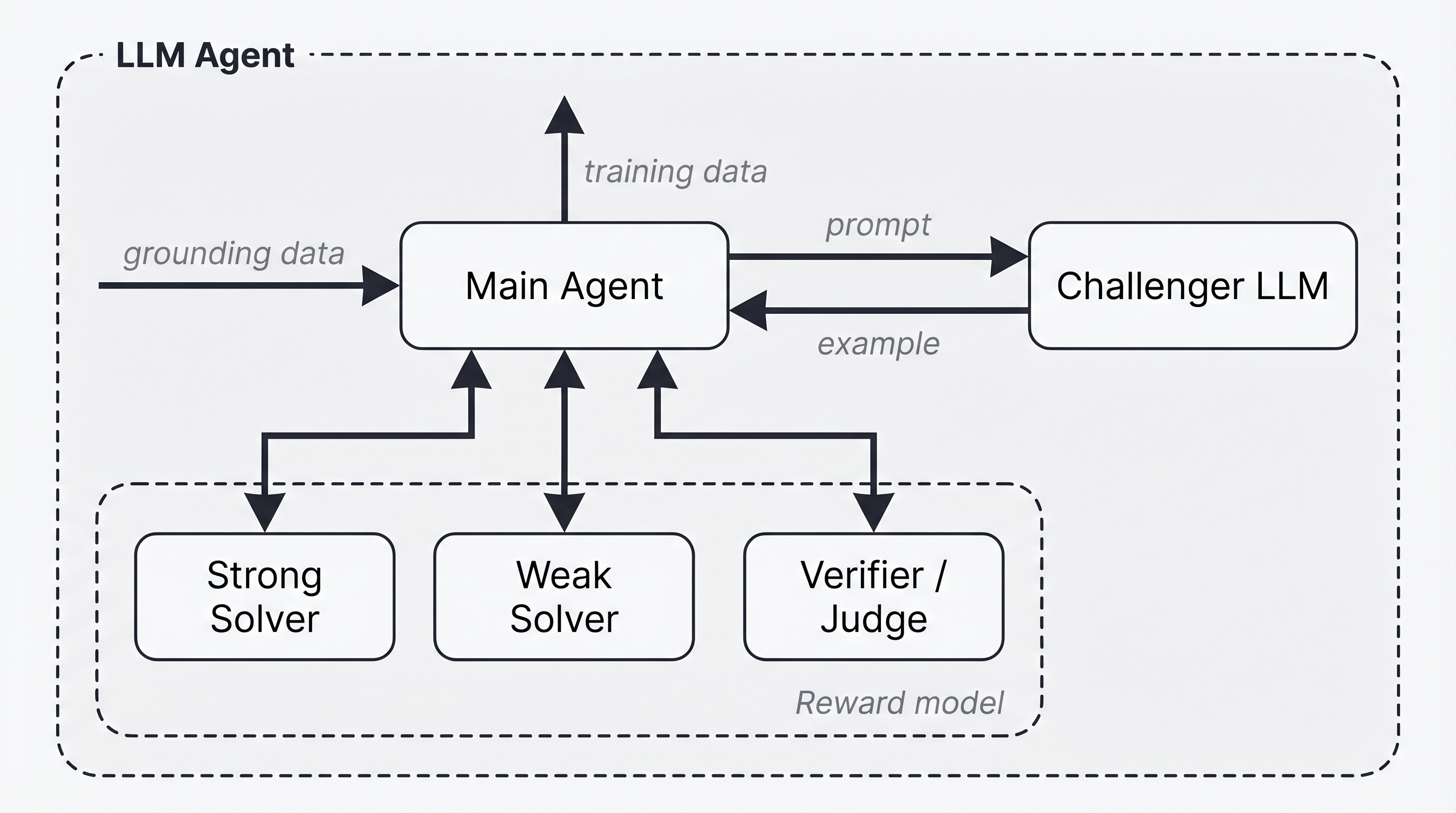}
  \caption{Weak-vs-strong {\bf Agentic Self-Instruct method}. The main LLM agent directs four subagents: a challenger LLM generates examples; weak and strong solvers attempt it; a judge evaluates their outputs. The system aims to generate training data where the strong solver succeeds while the weak solver struggles. The main LLM analyzes data and updates the challenger prompt using the judge’s feedback and repeats the cycle, yielding challenging examples for training the weak solver.}
  \label{fig:fig2}
\end{figure}

\subsection{A specific implementation: Agentic Self-Instruct}

In our experiments we consider a specific, practical implementation of Autodata for creating high quality data which we call Agentic Self-Instruct, depicted in 
  \autoref{fig:fig2}.
The main orchestrator agent has access to four LLM subagents: 
\begin{itemize}
\item[(i)] Challenger, which creates training examples given a detailed prompt from the main agent, 
\item[(ii)] ``Weak'' solver that is expected to generally struggle to solve the created training data; and 
\item[(iii)] ``Strong'' solver that is expected to generally succeed at the created training data,
\item[(iv)] Verifier/judge that given the example and a model solution, checks its quality, and passes its learnings back to the main agent.
\end{itemize}
The main agent proceeds to create an example (e.g., a given context/input, desired response or reference answer, and evaluation criteria depending on the task), by sending its initial prompt including grounding context to the challenger. It then checks the quality of the challenger's output by sending the input to the weak and strong solvers, and assigning a reward based on the verifier's judgments. 
The judge also plays the role of checking the quality of the example itself: the question, reference answer or generated rubric. 

For verifiable tasks (using an LLM-based verifier), one approach is to require that majority vote over the strong solver is correct, while majority vote over the weak solver is wrong. For non-verifiable tasks, we require a gap in quality as measured by the judge, e.g. given rubrics generated by the challenger, such that the task is neither too easy not too hard for the weak solver, while the strong solver helps guarantee correctness. The agent analyzes the report from the verifier (that includes the solver outputs), and if this criteria is not fulfilled, then it continues to modify the input prompt sent to the challenger given these new learnings, to try and make a new example until the criteria is met. 

This process allows the agent to effectively learn how to create challenging and high quality examples specifically for training the ``weak'' solver. We note that the ``weak'' and ``strong'' solvers can actually be the same LLM, but in different modes, e.g. the strong version can be allowed to use increased inference time compute including scaffolding or aggregation \citep{zhao2025majority}, as well as having access to privileged information.

\section{Experiments}

\subsection{Computer Science Research tasks}
\label{sec:cs}
We consider the task of answering computer science (CS) research questions, using academic CS papers as source material. %Unlike tasks such as mathematics where answers are verifiable through comparison with reference answers,
CS research questions are open-ended and, unlike verifiable tasks, require rubric-based evaluation. The challenger generates a context, a question, a reference answer, and a self-contained evaluation rubric consisting of weighted criteria that an LLM-judge uses to score any response without access to the reference answer.

\begin{table}[h]
  \centering
  \caption{Quality statistics for generated CS research tasks. CoT Self-Instruct is
  standard prompted generation; Agentic Self-Instruct is the agentic loop accepted output. Both columns are graded by Kimi-K2.6 at
  generation time on the same 4B-weak / 397B-strong solver pair.}
  \label{tab:cs_quality}
  \begin{tabular}{l c c}
  \toprule
  Metric & CoT Self-Instruct & Agentic Self-Instruct \\
  \midrule
  Weak solver avg        & 0.677 & 0.458 \\
  Strong solver avg      & 0.696 & 0.772 \\
  Gap (strong $-$ weak)  & 0.019 & 0.314 \\
  \midrule
  Agentic rounds      & 1.00   & {6.59}   \\
  Question length (chars) & 723    & {619}    \\
  Rubric items           & 13.2   & 13.1            \\
  \bottomrule
  \end{tabular}
\end{table}

\textbf{Pipeline overview.}
The main orchestrator agent calls the challenger to generate a context-QA pair with a corresponding rubric from a given paper. A quality verifier then checks for context leakage, rubric coverage, and question quality before final evaluation  (and also as a final step at the end of the loop). The question and context are sent to both solvers (each invoked 3 times to reduce variance), and the judge scores their answers against the rubric on a per-criterion basis. If any acceptance criterion fails, the agent provides targeted feedback to the challenger: which previous questions were too easy (with weak solver scores), which failed on the strong solver (with gap information), and which were rejected by the quality verifier. The challenger then generates a new question from a different reasoning angle. This loop typically runs several rounds per paper before it either finds an accepted question or exhausts the step budget.  
We use Kimi-K2.6 as the main orchestrator agent and challenger, Qwen3.5-397B-A17B as the strong solver, and Qwen3.5-4B as the weak solver. 
Agent system prompts are provided in Appendix
\autoref{app:prompts_cs}.

\textbf{Criteria.}
A useful training example for the weak solver requires that the strong solver scores meaningfully higher than the weak solver on the rubrics. In preliminary experiments, however, most questions generated by prompted Kimi-K2.6 are too \textit{easy} for the weak solver, leaving limited room for improvement: as shown in the ``CoT Self-Instruct'' column of \autoref{tab:cs_quality}, these questions have a weak solver average above 0.67 and a weak/strong gap of only 0.02. We therefore define the acceptance criterion of the agentic loop directly in terms of this gap: a candidate question is accepted only if the strong solver averages $\geq 0.65$, the weak solver $< 0.5$, and the strong$-$weak gap $\geq 20$ percentage points across the solver attempts. Given the strong nature of the weak solver, we save compute at each iteration by having the judge evaluate the strong solver only if the weak solver passes its corresponding success criterion. 

\textbf{Data Setup.}
We process over 10k CS papers from the S2ORC corpus (2022+) \citep{lo2020s2orc}, producing 2.8k accepted examples with Agentic Self-Instruct. The quality verifier with Kimi-K2.6 at the end of the loop (that removes questions with paper-specific reference leakage, short contexts, or malformed rubrics) further filters this set and we retain 1.3k high-quality accepted examples as the Agentic Self-Instruct data for RL training. For the CoT Self-Instruct baseline we also apply the same quality verifier and sample an equal amount of 1.3k filtered data for fair comparison.

\begin{table}[t]
  \centering
  \caption{
  RL training results on CS research tasks. The autodata Agentic Self-Instruct method outperforms creating data with standard CoT Self-Instruct.
  We train Qwen3.5-4B with GRPO
  on 1.3k examples from each data source and evaluate on a 200-prompt held-out test
  set. Scores are rubric-based, graded by Kimi-K2.6. As shown in
  \autoref{fig:cs_training_curves}, Agentic outperforms CoT data throughout training;
  we illustrate with step 200 in the table.}
  \label{tab:cs_rl}
  \begin{tabular}{l cc cc}
  \toprule
   & \multicolumn{2}{c}{\textbf{CoT test}} & \multicolumn{2}{c}{\textbf{Agentic test}} \\
  \cmidrule(lr){2-3} \cmidrule(lr){4-5}
  Response model & mean@3 & best@3 & mean@3 & best@3 \\
  \midrule
  Qwen3.5-4B (no additional RL)                              & 0.630 & 0.758 & 0.366 & 0.484 \\
  Qwen3.5-4B RL on CoT Self-Instruct  data             & 0.727 & 0.853 & 0.500 & 0.631 \\
  \textbf{Qwen3.5-4B RL on Agentic Self-Instruct data} & \textbf{0.774} & \textbf{0.894} &
  \textbf{0.632} & \textbf{0.768} \\
  \bottomrule
  \end{tabular}
\end{table}

\begin{figure}[t]
  \centering
  \includegraphics[width=\textwidth]{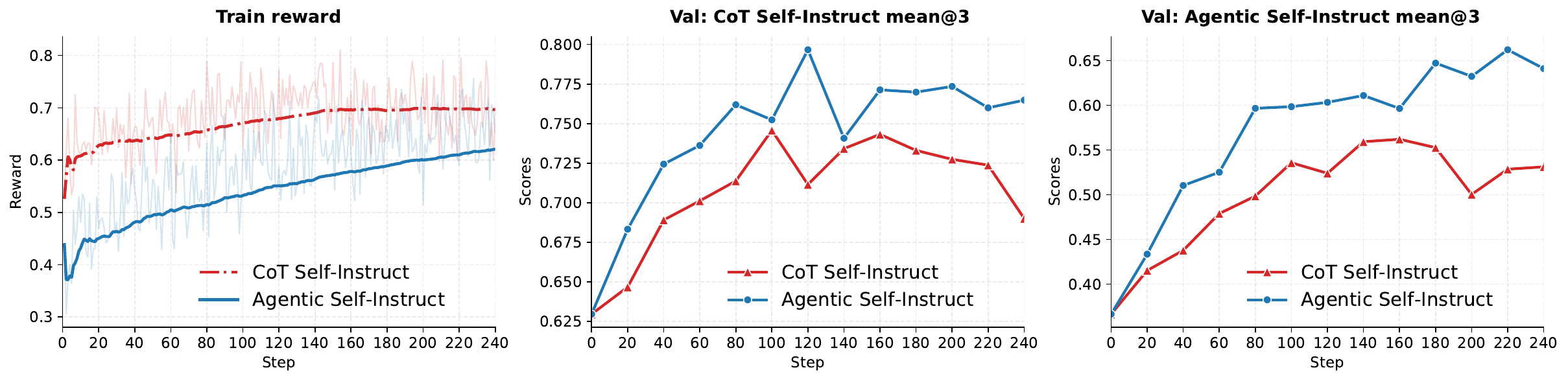}
  \caption{Train reward and held-out validation for the two CS RL runs
  (Qwen3.5-4B trained on Agentic vs.\ CoT Self-Instruct data). \textit{Left}:
  per-step Kimi-K2.6 rubric reward, EMA-smoothed. \textit{Middle / Right}:
  mean@3 on the CoT and Agentic held-out test sets (each is in-distribution
  for one arm, out-of-distribution for the other). Agentic-trained leads on both held-out sets throughout training, with the larger margin on the harder Agentic test.}
  \label{fig:cs_training_curves}
\end{figure}

\subsubsection{Agentic Self-Instruct Loop Analysis}

\label{sec:cs_loop}

Running the data creation loop over our 10k-paper corpus, the agent needed substantial iterations to produce an accepted question: a mean of 6.59  rounds per accepted item (\autoref{tab:cs_quality}), with a long tail extending past 10 rounds on a fraction of papers. The failure modes we saw at each round were heavily one-sided: across 880 pre-acceptance rounds, 80\% of failed rounds were rejected because the question was too easy and the weak solver scored too high, 13\% because the strong solver could not reliably solve them either.

For the papers that did converge, the generated question was rarely the one that the agent created in the first attempt. For instance, the weak solver scores an average of 0.677 on the questions generated by the baseline CoT Self-Instruct method, while the questions generated using Agentic Self-Instruct see a 22-point drop given the same source material (papers). Inspecting the trajectories, we observe that the agent's initial attempt on a CS paper was usually a high-level summary question that often proves to be easy for a 4B solver model. However, subsequent rounds, guided by the judge's feedback, moved the questions toward specific algorithmic steps, ablation details, or numerical claims in the paper that required following the paper's actual argument. The aggregate effect of this search is visible in the corpus statistics in \autoref{tab:cs_quality}: the weak solver's score drops by 22 points (0.677 $\to$ 0.458) while the strong solver's score improves by 8 points (0.696 $\to$ 0.772), confirming that the accepted questions are harder in a way the strong solver's deeper reasoning specifically rewards. That is, the agentic data creation loop produces questions that specifically reward stronger model capabilities, rather than questions both models can answer.

\begin{figure}
    \centering
    \includegraphics[width=1\linewidth]{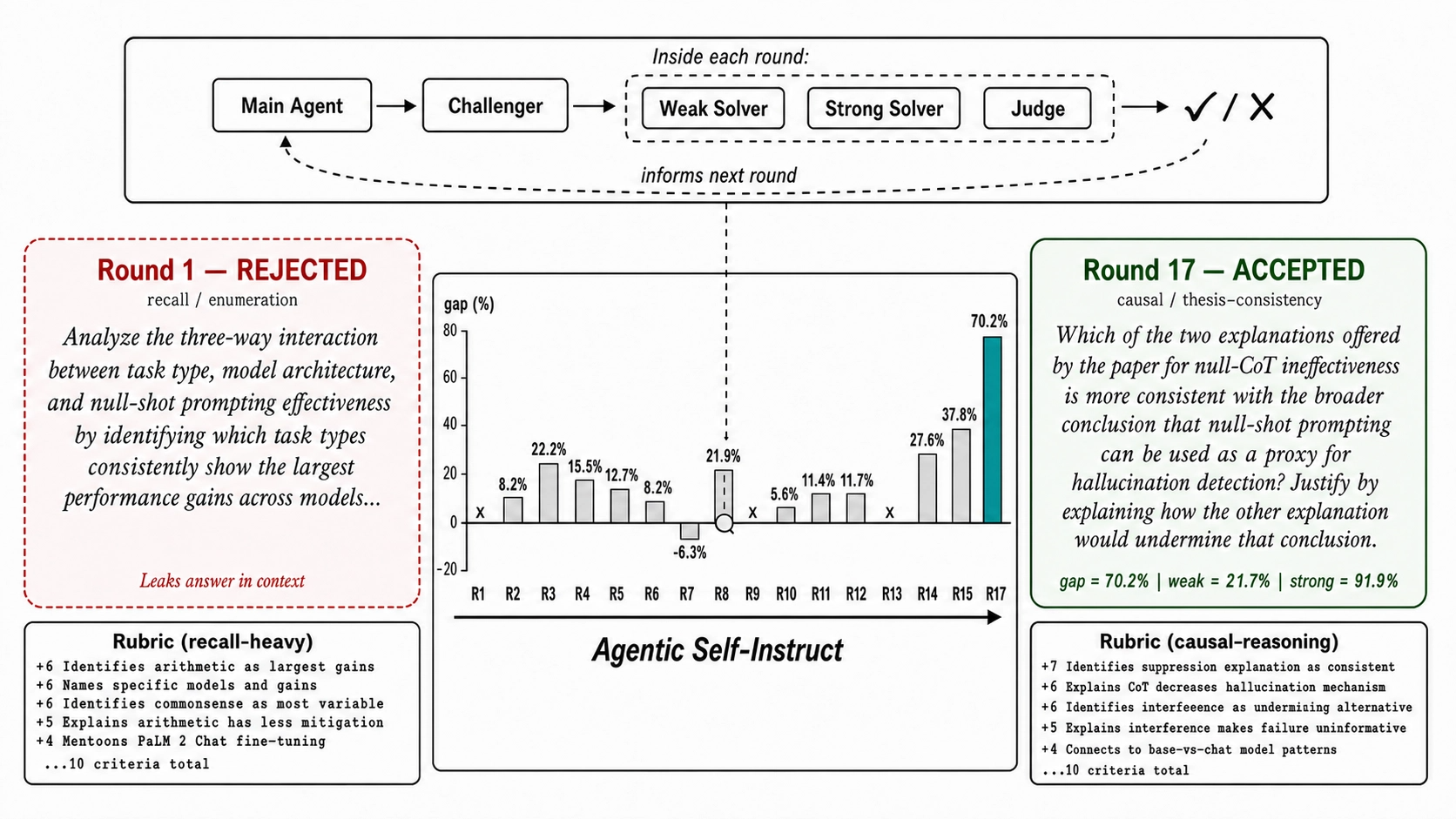}
    \caption{Autodata creation of CS research questions. Shown is the progression of Agentic Self-Instruct generating a training example with corresponding evaluation rubric for a given paper about Large Language Modeling. }
    \label{fig:cs_round6_example}
\end{figure}

Improvement works through exploration. Each round the agent generates a new question from a different reasoning angle, guided by feedback on which previous questions were too easy or failed to discriminate. The accepted questions after the agentic loop also qualitatively test different reasoning types: specific technical mechanisms, multi-step derivations, and paper-specific design tradeoffs, compared to the broader, more generic questions produced without this loop. See \autoref{fig:cs_round6_example} for an example.

\subsubsection{RL Training Results}

We compare the performance of Qwen3.5-4B trained on 1.3k examples from CoT Self-Instruct versus Agentic Self-Instruct data, using Kimi-K2.6 as the reward model to score responses against the generated rubrics. From each dataset (generated using CoT Self-Instruct and Agentic Self-Instruct) we use 100 examples as the test set and train Qwen3.5-4B with GRPO \citep{shao2024deepseekmath} (batch size 16, learning rate 1e-6), evaluating each trained model on both held-out test sets. 

On the easier CoT Self-Instruct test set (\autoref{tab:cs_rl}, left), training on CoT data lifts the base 4B model from 0.630 mean@3 to 0.727 and training on Agentic data lifts it further to 0.774. On the harder Agentic test set (right), the corresponding numbers are 0.366 (base) $\to$ 0.500 (CoT-trained) $\to$ 0.632 (Agentic-trained): the gap between the two methods is more than twice as large here as on the CoT test, and best@3 follows the same ordering. The training dynamics in \autoref{fig:cs_training_curves} are consistent with this: our Agentic method sits above the CoT Self-Instruct method on the per-step Kimi reward from the start and the spread widens through training, while on the validation panels our Agentic method matches or exceeds the CoT method at every checkpoint on both test sets, including the one where the CoT data was the natural in-distribution choice. The Agentic-trained model transfers in both directions ($+0.05$ to the easier CoT test, $+0.13$ to the harder Agentic test), the clear advantage suggests that the discriminative training data produced by the Agentic pipeline translates to stronger reasoning performance.

\subsection{Legal Reasoning Tasks}
\label{sec:legal}

We next investigate a second setting to test the method's generality, on legal reasoning tasks.
In \autoref{sec:cs} we experimented with Agentic Self-Instruct on open-ended CS research tasks, where the initial synthetic data was not challenging enough for the model to improve when using it for downstream RL training. This section studies a second setting on improving an LLM's legal reasoning capabilities, which turns out to have different qualities. Here, we find that Agentic Self-Instruct has to contend with the opposite failure mode: we find standard prompting via CoT self-instruct produces questions that are \emph{too hard} (and not \emph{too easy}) in a way that hinders the RL reward signal.

The goal as before is to create high-quality data to improve the weak solver on the given task, in this case legal reasoning. We use court opinions and other public legal documents drawn from Pile of Law~\citep{henderson2022pile} as source material, and evaluate on PRBench-Legal and the PRBench-Legal-Hard subset \citep{akyurek2025prbench}. As in the CS setting, we use Kimi-K2.6 as the main orchestrator agent, challenger and judge, Qwen3.5-397B-A17B as the strong solver, and Qwen3.5-4B as the weak solver. Quite differently from CS papers, our initial study showed that CoT Self-Instruct generated questions and rubrics are too hard for the weak solver. As shown in \autoref{tab:legal_quality}, the weak solver averages only 0.159, with many of the attempts scoring 0, which hinders learning under GRPO. If we were to apply the hard-threshold acceptance criteria from the CS setting here, the majority of the CoT data points would have been accepted. Instead, we ask: can we directly verify both the quality and the GRPO-suitability of a data point?

\begin{table}[h]
  \centering
  \small
  \caption{Quality statistics for generated legal reasoning tasks.
  CoT Self-Instruct is standard prompted generation; Agentic Self-Instruct is the output of the agentic loop. Both columns are graded by Kimi-K2.6
  at generation time on the same 4B-weak / 397B-strong solver pair.}
  \label{tab:legal_quality}
  \begin{tabular}{l c c}
  \toprule
  Metric & CoT Self-Instruct  & Agentic Self-Instruct  \\
  \midrule
  Weak solver avg      & 0.159 & 0.283 \\
  Strong solver avg      & 0.717 & 0.698          \\
  Gap (strong $-$ weak)  & 0.558 & 0.415 \\
  \midrule
  Agentic rounds      & 1.00   & {4.98}   \\
  Question length (chars)& 1{,}569 & {900}   \\
  Rubric items           & 18.6   & 17.3            \\
  Weak rollout std       & 7.93   & {12.63}  \\
  \bottomrule
  \end{tabular}
\end{table}

\textbf{Pipeline overview.}  The main difference from the hard-coded acceptance criteria we adopted  in the CS papers task is that here we instead adopt a more flexible {\em loop judge} to decide if a round's generation is accepted. Specifically, each legal document is first passed through a dedicated \emph{extractor} agent that produces a structured extract (topic keywords, salient facts, holdings). The \emph{challenger} agent then generates one realistic legal question plus a weighted grading rubric and a declared target-capability set from the extract. Each candidate is rolled out by the weak solver 5 times and the strong solver 3 times. 

The judge then reads the per-rollout solver patterns, the weak/strong gap, and the rubric and returns a structured verdict (\texttt{weak\_pattern}, \texttt{strong\_pattern}, \texttt{gap\_interpretation}, \texttt{rubric\_concerns}, \texttt{grpo\_suitability}) plus an \texttt{accept}/\texttt{improve} decision. For the \texttt{improve} decision case, it hands the challenger a concrete \texttt{suggestion\_for\_challenger} (e.g.\ ``the weak-rollouts are all reciting the same boilerplate; push the question toward step-wise application of the holding rather than recall''), and the loop re-runs; on \texttt{accept} the example is considered good quality and the loop ends. Unlike the hard-coded CS quality decision, the judge we use for legal tasks has no fixed acceptance thresholds: it decides per round given analysis of the results, including the weak-rollout \emph{variance}, the gap, and PRBench baseline solver statistics. A good legal training example is defined by overall data quality and GRPO-suitability, not only by a numeric gap target. 

Agent system prompts are provided in Appendix \autoref{app:prompts_legal}.

\textbf{Data Setup.} We processed 7.8k source documents, where 5.7k of these produced usable  CoT Self-Instruct examples and 2.8k reached an \texttt{accept} verdict after the agentic loop. For the controlled head-to-head RL comparison in \autoref{sec:legal_rl}, the \emph{Agentic} experiment uses all 2.8k accepted data points and the \emph{CoT} Self-Instruct set uses 2.8k examples randomly drawn from the 5.7k CoT pool.

\begin{table}[t]
  \centering
  \caption{RL training results evaluated on PRBench.  The autodata Agentic Self-Instruct method outperforms creating data with standard CoT Self-Instruct.
  We train Qwen3.5-4B with GRPO on 2.8k
    legal QA pairs from each data source and evaluate on
    the PRBench-Legal (500 prompts) and PRBench-Legal-Hard (250-prompt subset)
    test sets. Scores are clipped per-prompt PRBench scores,
    graded by both Kimi-K2.6 and GPT-5. Both graders agree: our 4B trained on
    Agentic data outperforms the model trained on the standard CoT Self-Instruct data as well as the much larger 397B baseline on both splits.}
  \label{tab:prbench_headline_E}
  \begin{tabular}{l cc cc}
  \toprule
   & \multicolumn{2}{c}{\textbf{GPT-5 Grader}} & \multicolumn{2}{c}{\textbf{Kimi-K2.6 Grader}}
   \\
  \cmidrule(lr){2-3} \cmidrule(lr){4-5}
  Response Model & Legal & Legal-Hard & Legal & Legal-Hard \\
  \midrule
  Qwen3.5-4B (no additional RL)                              & 0.280 & 0.167 & 0.245 & 0.145 \\
  Qwen3.5-397B (no additional RL)                            & 0.404 & 0.277 & 0.358 & 0.226 \\
  \midrule
  Qwen3.5-4B RL on CoT Self-Instruct data              & 0.377 & 0.253 & 0.343 & 0.233 \\
  \textbf{Qwen3.5-4B RL on Agentic Self-Instruct data} & \textbf{0.441} & \textbf{0.315} &
  \textbf{0.393} & \textbf{0.266} \\
  \bottomrule
  \end{tabular}
  \end{table}
  
\begin{figure}[t]
  \centering
  \includegraphics[width=\textwidth]{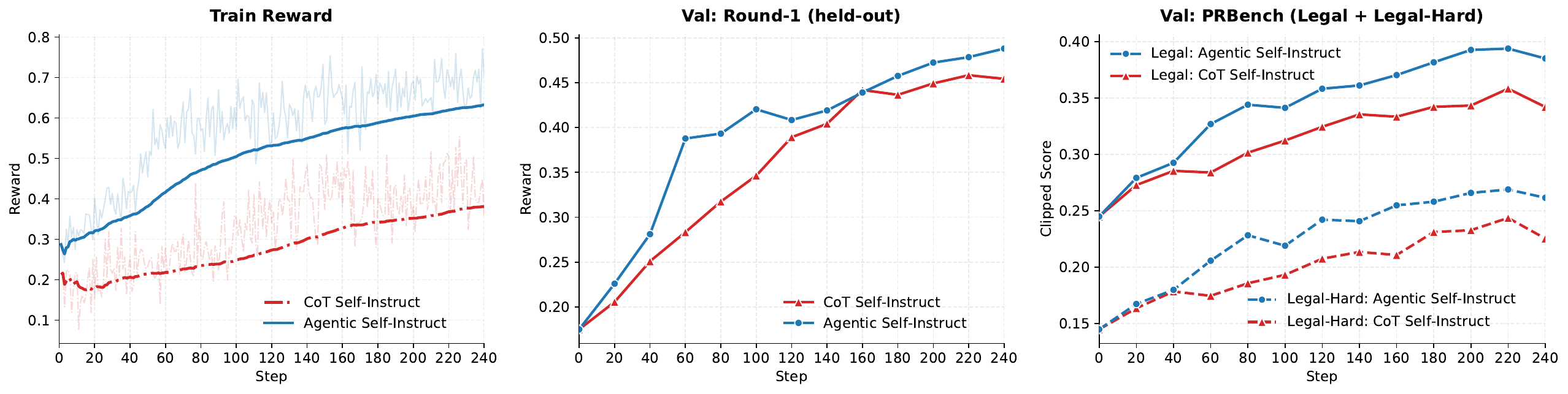}
  \caption{RL training dynamics on legal reasoning. We train Qwen3.5-4B with GRPO on
  2.8k legal Question-Rubric pairs from each data source (Agentic
  Self-Instruct, CoT Self-Instruct) and evaluate every 20 steps on a
  100-prompt held-out CoT set (\textit{middle}) and the PRBench Legal /
  Legal-Hard splits (\textit{right}). All rewards and scores graded by
  Kimi-K2.6. Agentic stays ahead of CoT on every metric throughout training.}
  \label{fig:training_curves_2804}
\end{figure}

\subsubsection{Agentic Self-Instruct Loop Analysis}
\label{sec:legal_loop}
\paragraph{The Agentic Self-Instruct loop reshapes the weak-rollout distribution.}
\autoref{tab:legal_quality} shows the before/after analysis: the weak/strong gap on the same source documents actually narrows from 55.8 to 41.5 points after the agentic loop, unlike the CS setting. The notable change lies in that per-prompt weak-rollout standard deviation rises from 7.93 to 12.63. CoT questions concentrate weak scores near zero (mean 15.9\%, median 10.7\%); many prompts have four or five out of five weak-rollouts scoring zero, which leaves the per-group GRPO advantage near zero and provides little learning signal. The agentic loop pushes the weak mean up to 28.3\% while leaving strong roughly unchanged (71.7\% $\to$ 69.8\%); the same gap is now spread over a usable variance range. The loop makes the questions more \emph{learnable} by reshaping the per-prompt reward signal. As a byproduct, the loop judge's textual feedback also pushes the challenger toward shorter, application-style questions (mean 900 vs 1.6k characters) without changing rubric density, incidentally aligning the format with the relatively short PRBench-Legal prompts.

At each round the loop judge provides a categorical \texttt{grpo\_suitability} verdict (\texttt{high}/\texttt{medium}/\texttt{low}) based on the rollout patterns. On the CoT Self-Instruct pool the distribution is 4.8\% high / 41\% medium / 45\% low; on the Agentic pool it is \textbf{52\% high / 43\% medium / 2\% low}. The median accepted question goes through 4 agentic rounds (mean 4.98, max 19), only $\sim$2\% use a single round.

\subsubsection{RL Training Results}
\label{sec:legal_rl}

We RL-train Qwen3.5-4B with GRPO on the setups: 2.8k Agentic Self-Instruct prompts (\emph{Agentic}) versus 2.8k standard CoT  Self-Instruct prompts (\emph{CoT}), with $n{=}8$ rollouts per prompt, and a Kimi-K2.6 rubric judge as the train-time reward. We evaluate on PRBench-Legal (500 prompts, 250 of them tagged \texttt{legal\_hard}) under the PRBench-official judge setting (e.g, the judge evaluates each rubric separately), and additionally re-grade the same rollouts with GPT-5 as an independent, stronger grader to confirm the comparison is not Kimi-grader biased.

On the 500-prompt PRBench-Legal split (\autoref{tab:prbench_headline_E}), Qwen3.5-4B RL'd on Agentic data scores 0.441 (GPT-5 as judge) and 0.393 (Kimi as judge), outperforming the same-architecture CoT-trained model (0.377 / 0.343), and even outperforming the much larger strong Qwen3.5-397B-A17B baseline without additional RL (0.404 / 0.358). The same ordering holds on PRBench-Legal-Hard. The $+0.05$--$0.06$ advantage of Agentic over CoT is obtained on the same 2.8k-prompt budget, same challenger, same source corpus: the only difference between the two setups is the agentic loop in the training data creation. The training curves in \autoref{fig:training_curves_2804} show the Agentic method leading on train reward, on the held-out CoT validation set, and on PRBench-Legal at every checkpoint we evaluated.

\subsubsection*{``More Challenging'' vs ``Just Right''}

The two tasks we have so far experimented with in \autoref{sec:cs} and \autoref{sec:legal} apply the \emph{same} Agentic Self-Instruct loop to opposite failure modes of standard CoT-Self Instruct prompt-based generation: in CS, CoT questions are too easy for the weak solver (gap 0.02, raising the concern that the questions are not challenging enough), while in Legal they are too hard, with many rollouts scored at 0 (gap 0.56, but providing too harsh a learning signal). After applying the Autodata agentic loop the gap moves in opposite directions (widening in CS, narrowing in Legal), yet the downstream RL outcome is the same: the model trained on autodata-generated data outperforms the model trained on CoT data on every held-out test, and on PRBench-Legal a 4B model outperforms a much larger baseline. 
The key is not to make the question more challenging, but to make them \emph{just right} for the model to hill-climb on; the Agentic Self-Instruct loop is what lets us achieve this.

\subsection{Scientific reasoning}

Next, we consider the construction of challenging problems that require reasoning over mathematical objects in the same categories and domains as the existing Principia collection \citep{aggarwal2026reasoning}. The Principia collection was designed using a CoT Self-Instruct-based method (prompted, multi-step LLM workflow) covering a wide range of curricula from the MSC2020 and PHYS catalogs. Principia bench, on the other hand, consists of human-labeled subsets of existing math and physics benchmarks where the problems were filtered to contain those involving mathematical objects in the answer.

\textbf{Pipeline overview} In this experiment, we use a weak solver of Qwen3.5-4B, and a strong solver of Qwen3.5-397B-A17B. The weak solver is in fact a very capable reasoning model that can solve many problems from the Principia collection. Thus it is a good candidate model for our pipeline where we seek to generate problems that are more challenging for the weak solver.
The main orchestrator agent and challenger are Kimi K2.6. Agent system prompts are provided in Appendix
\autoref{app:prompts_math}.
A detailed breakdown of question types in the constructed agentic data is provided in~\autoref{app:question_types}.

\subsubsection{RL training results}

This setup allows us to compare three data sources for down-stream RL training: (i) \textbf{CoT Self-Instruct}: training directly on the problems from the Principia collection, which are also used as grounding context in Agentic Self-Instruct, (ii) \textbf{Agentic}: training on data generated by  Agentic Self-Instruct, and (iii) \textbf{Combined}: training on both data sources together, which doubles the training set size. Each individual data source consists of 9k training examples and 1k held-out evaluation examples, while the Combined setting uses 18k training examples. Similarly to previous experiments, we train Qwen3.5-4B model using Kimi K2.6 as a judge to compare model's generated answer against the reference answer and assign binary reward based on the comparison. We use GRPO with group size 8 and batch size 64 for training.

We evaluate models both in and out of their training distributions using a combined validation set consisting of held-out examples from both the agentic-generated and CoT data distributions, as well as the out-of-distribution Principia benchmark. Results are shown in \autoref{tab:principia_grounded_combined_tabs} and \autoref{tab:principia_grounded_bench_tabs}.

\begin{table}[t]
  \centering
  \caption{RL training results evaluated on scientific reasoning tasks. Agentic Self-Instruct data outperforms CoT Self-Instruct or even Combined data ($2\times$ training size). Deltas ($\Delta$) are relative to the starting Qwen3.5-4B model.}
  \label{tab:principia_grounded_combined_tabs}
  \resizebox{0.85\textwidth}{!}{%
  \begin{tabular}{l c cc cc cc}
  \toprule
   & \textbf{Base} & \multicolumn{2}{c}{\textbf{CoT Self-Instruct}} & \multicolumn{2}{c}{\textbf{Agentic Self-Instruct}} & \multicolumn{2}{c}{\textbf{Combined ($2\times$ data)}} \\
  \cmidrule(lr){3-4} \cmidrule(lr){5-6} \cmidrule(lr){7-8}
  \textbf{Eval Subset} & Qwen3.5-4B & avg@8 & $\Delta$ & avg@8 & $\Delta$ & avg@8 & $\Delta$ \\
  \midrule
  \textbf{Overall} & 68.66\% & 71.08\% & +2.42 & \textbf{71.86\%} & \textbf{+3.20} & 71.36\% & +2.70 \\
  Agentic subset & 52.39\% & 56.33\% & +3.94 & \textbf{56.79\%} & \textbf{+4.40} & 55.88\% & +3.49 \\
  CoT Self-Instruct subset & 77.17\% & 79.03\% & +1.86 & \textbf{80.22\%} & \textbf{+3.05} & 79.66\% & +2.49 \\
  \midrule
   & & pass@8 & $\Delta$ & pass@8 & $\Delta$ & pass@8 & $\Delta$ \\
  \midrule
  \textbf{Overall} & 87.73\% & 88.58\% & +0.85 & \textbf{88.91\%} & \textbf{+1.18} & 88.75\% & +1.02 \\
  Agentic subset & 81.13\% & 81.74\% & +0.61 & 81.86\% & +0.73 & \textbf{82.11\%} & \textbf{+0.98} \\
  CoT Self-Instruct subset & 92.52\% & 92.77\% & +0.25 & \textbf{94.36\%} & \textbf{+1.84} & 93.50\% & +0.98 \\
  \bottomrule
  \end{tabular}%
  }
\end{table}

\begin{table}[t]
  \centering
  \caption{Out-of-distribution Principia benchmark results comparing training data sources. Agentic Self-Instruct data yields the largest overall improvement despite using half the data of Combined.}
  \label{tab:principia_grounded_bench_tabs}
  \resizebox{0.85\textwidth}{!}{%
  \begin{tabular}{l c c cc cc cc}
  \toprule
   & & \textbf{Base} & \multicolumn{2}{c}{\textbf{CoT Self-Instruct}} & \multicolumn{2}{c}{\textbf{Agentic Self-Instruct}} & \multicolumn{2}{c}{\textbf{Combined ($2\times$ data)}} \\
  \cmidrule(lr){4-5} \cmidrule(lr){6-7} \cmidrule(lr){8-9}
  \textbf{Category} & \textbf{Items} & Qwen3.5-4B & avg@8 & $\Delta$ & avg@8 & $\Delta$ & avg@8 & $\Delta$ \\
  \midrule
  \textbf{Overall} & 2113 & 50.43\% & 51.10\% & +0.67 & \textbf{51.47\%} & \textbf{+1.04} & 51.17\% & +0.74 \\
  ARB & 47 & 81.91\% & \textbf{83.78\%} & \textbf{+1.87} & 80.32\% & $-$1.59 & 83.24\% & +1.33 \\
  Physics & 110 & 66.22\% & 65.91\% & $-$0.31 & \textbf{67.05\%} & \textbf{+0.83} & 66.25\% & +0.03 \\
  RealMath & 632 & 33.68\% & 35.25\% & +1.57 & \textbf{35.43\%} & \textbf{+1.75} & 34.97\% & +1.29 \\
  SuperGPQA & 1324 & 56.00\% & 56.28\% & +0.28 & \textbf{56.82\%} & \textbf{+0.82} & 56.50\% & +0.50 \\
  \midrule
   & & & pass@8 & $\Delta$ & pass@8 & $\Delta$ & pass@8 & $\Delta$ \\
  \midrule
  \textbf{Overall} & 2113 & 59.54\% & \textbf{60.06\%} & \textbf{+0.52} & 59.91\% & +0.37 & 59.91\% & +0.37 \\
  ARB & 47 & 91.49\% & 91.49\% & +0.00 & 91.49\% & +0.00 & \textbf{93.62\%} & \textbf{+2.13} \\
  Physics & 110 & 76.36\% & 77.27\% & +0.91 & \textbf{79.09\%} & \textbf{+2.73} & 72.73\% & $-$3.63 \\
  RealMath & 632 & 41.93\% & 43.67\% & +1.74 & 43.67\% & +1.74 & \textbf{44.30\%} & \textbf{+2.37} \\
  SuperGPQA & 1324 & 65.41\% & 65.33\% & $-$0.08 & 64.95\% & $-$0.46 & 65.11\% & $-$0.30 \\
  \bottomrule
  \end{tabular}%
  }
\end{table}

On the combined validation set (\autoref{tab:principia_grounded_combined_tabs}), training on Agentic Self-Instruct data yields the largest overall improvement (+3.20\% avg@8), outperforming both direct training on CoT Self-Instruct data (+2.42\%) and the combined dataset (+2.70\%). A key finding is that Agentic Self-Instruct data improves performance even on the CoT validaton subset (+3.05\% vs.\ +1.86\% for CoT Self-Instruct), despite not being explicitly optimized for that distribution. This demonstrates that \emph{training on harder problems transfers to easier ones}: the challenging examples produced by our iterative agentic process teach reasoning skills that generalize beyond the specific difficulty level they target.

On the out-of-distribution Principia benchmark (\autoref{tab:principia_grounded_bench_tabs}), Agentic Self-Instruct again achieves the best overall avg@8 improvement (+1.04\%), with consistent gains across most categories, particularly on RealMath (+1.75\%) and SuperGPQA (+0.82\%). This transfer effect further confirms that the harder problems generated by Agentic Self-Instruct build more robust reasoning capabilities.

The pass@8 results reveal a more nuanced picture with trade-offs across methods. The Combined data shows advantages on pass@8 in several categories: ARB (+2.13\% vs.\ +0.00\% for both Agentic and Grounding) and RealMath (+2.37\% vs.\ +1.74\% for Agentic). This suggests that while Agentic Self-Instruct improves \emph{average} performance by teaching the model to solve challenging problems more reliably, the Combined data's greater diversity (and size) may help the model \emph{occasionally} solve a broader range of problems (reflected in higher pass@8). One possible explanation is that Qwen3.5-4B may be approaching its capacity limit for this task distribution, and larger models might better exploit the combined data to achieve gains on both metrics simultaneously.

These results highlight the value of \emph{data quality and difficulty}. The iterative process in Agentic Self-Instruct produces challenging examples that provide a more efficient learning signal---training on these harder problems not only improves performance on difficult tasks but also transfers to easier ones. This supports the hypothesis that investing inference-time compute in generating higher-quality, more challenging synthetic data can be more effective than simply scaling dataset size. Additionally, we find that training substantially reduces reasoning truncation rates (from 23.75\% to 4.09\% for Agentic Self-Instruct with a 65,536 token budget), with approximately half of accuracy improvements attributable to more token-efficient reasoning (see \autoref{app:truncation}).

\section{Meta Optimization of the Data Scientist}

So far, we have implemented the autodata agent using a fixed Agentic Self-Instruct framework, with provided prompts which define how the agent's strategy. However, it is also possible to (meta-)learn the overall agent as well. In this section we thus apply meta-optimization to the data scientist agent itself, using the same evaluation criteria from the inner loop to guide optimization of the outer loop---the agent's prompt and strategy. Concretely, we use a evolution optimization framework that treats the agent's scaffold as code to be iteratively improved.

\begin{figure}[tbp!]
  \centering
  \includegraphics[width=1.01\linewidth]{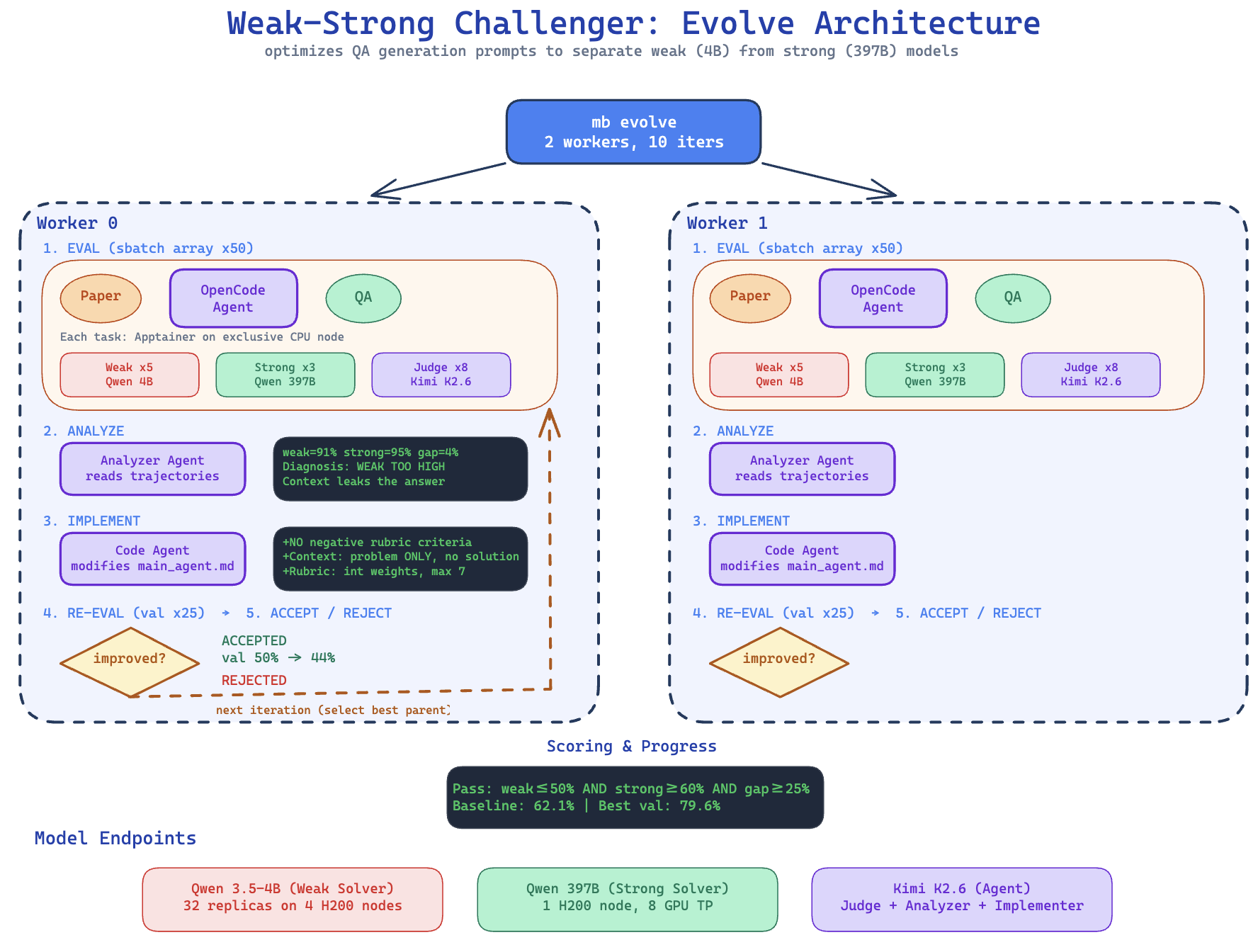}
 \caption{Meta-optimization of the data scientist agent. An outer optimization loop evaluates the agent's prompt on training examples of CS paper task, analyzes failure trajectories to identify systematic weaknesses (e.g., context leakage), implements prompt modifications via a code-editing agent, and re-evaluates on held-out validation papers. Changes are accepted only if they improve the weak-strong separation rate. This process improved validation pass rate from 62.1\% to 79.6\% over 126 accepted iterations out of 233 total.}

  \label{fig:meta_train}
\end{figure}

\paragraph{Method.} The meta-optimizer maintains a population of candidate prompts, each defined by a code diff relative to the baseline repository. Each iteration proceeds as follows: (1)~\textbf{Select} a parent from the population via Boltzmann sampling, where candidate $c$ is chosen with probability proportional to $\exp(score_c / T)$ with temperature $T{=}0.1$, strongly favoring high-scoring candidates while maintaining exploration; (2)~\textbf{Evaluate} the parent's prompt on a minibatch of training papers, collecting agent trajectories and weak/strong solver scores; (3)~\textbf{Analyze} the trajectories with an LLM agent that reads the full solver exchanges and writes a root-cause analysis of systematic failure patterns; (4)~\textbf{Implement} prompt modifications via a code-editing agent that reads the analysis, iteration history, and current prompt, then produces an improved diff; (5)~\textbf{Re-evaluate} both parent and mutant on held-out validation papers; (6)~\textbf{Accept or reject} the mutant---it is added to the population only if its validation score strictly exceeds its parent's; (7)~\textbf{Summarize} the outcome into a history log that subsequent analyzers can read. Multiple iterations run concurrently with independent parent selections. Because single-evaluation scores are noisy (solvers run at temperature 1.0), accepted candidates accumulate additional evaluations whenever they are subsequently sampled as parents, and we report the averaged score across all re-evaluations.

\paragraph{Setup.} We meta-optimize the CS research paper task from \autoref{sec:cs}. The meta-optimizer uses Kimi-K2.6 as both the analyzer (which reads evaluation trajectories to diagnose failure patterns) and the implementer (which modifies the agent's prompts). The inner-loop agent being optimized also uses Kimi-K2.6 in a multi-agent configuration with separate challenger, main agent, and quality verifier prompts, following \autoref{sec:cs}. We use 50 training papers and 25 validation papers. A generated QA pair is considered successful if it satisfy all of the criterion's: weak solver (Qwen3.5-4B) scores $\leq$65\%, the best weak solver attempt score $\leq 75\%$, the strong solver (Qwen3.5-397B-A17B) scores $\geq$60\% and $\leq$95\%, and the gap between the strong and weak solver $\geq$20 percentage points, as judged by rubric-based evaluation.

\paragraph{Results.} Starting from a baseline prompt that achieves $\sim$62.1\% validation QA pass rate on the 100 validation samples, the meta-optimizer progressively discovers prompt improvements across 233 iterations. We report the baseline and the final evolved harness result under 6h per session timeout. Results are shown in \autoref{tab:meta_opt_results}.

\begin{table}[h]
\centering
\caption{Meta-optimization of the data scientist agent on the CS research paper task. The optimizer iteratively improves the agent's prompt, with each accepted iteration building on the previous best. Validation pass rate (re-evaluated) measures the fraction of generated QA pairs that successfully separate weak and strong solvers, averaged over multiple re-evaluations to reduce noise.}
\begin{tabular}{lcc}
\toprule
\textbf{Meta Iteration} & \textbf{Num Val Sample} & \textbf{Val Pass Rate 6h} \\
\midrule
 Baseline & 100 & 62.1\% \\
 Iter 124 & 100 & 79.6\% \\
\bottomrule
\end{tabular}
\label{tab:meta_opt_results}
\end{table}

The meta-optimizer identified several systematic failure modes through trajectory analysis---examining what the weak solver actually said in its responses and identifying that generic answers and rubric format errors were the dominant causes of poor separation. The optimizer addressed these through the following prompt modifications, discovered automatically over the course of 233 iterations:

\begin{itemize}[nosep]
\item \textbf{Paper-specific insight enforcement}: The optimizer added instructions requiring that questions test knowledge \emph{specific to the paper}, not generic ML/CS knowledge. A self-test was introduced: ``If a solver could answer correctly without reading this specific paper, the question is too easy.'' This directly addressed weak solvers achieving high scores by producing plausible-sounding generic responses.
\item \textbf{Context leak prevention}: Strict rules were added requiring the context to describe only the problem domain and setup, never the paper's proposed solution. A self-test was introduced: ``Could someone answer the question by rephrasing sentences from the context? If yes, rewrite.''
\item \textbf{Positive-only rubric with weight capping}: The optimizer \emph{eliminated} negative-weight rubric criteria, finding that they historically misfired and destroyed strong model scores without improving discrimination. Instead, all criteria use positive integer weights capped at 7, preventing any single criterion from dominating the score. This was a counter-intuitive discovery---penalizing errors seemed helpful in theory but hurt in practice.
\item \textbf{Structured rubric format}: The optimizer enforced a strict JSON format for rubric criteria with integer weights, eliminating parsing errors (e.g., string weights like ``+8'' instead of the integer 8) that had caused evaluation failures in earlier iterations.
\end{itemize}

The progression from 62.1\% to 79.6\% validated pass rate demonstrates that meta-optimizing the data scientist agent's instructions can substantially improve data quality without manual prompt engineering, though the absolute numbers also highlight the difficulty of reliably generating questions that separate models of different capability levels.

\section{Related Work}

\paragraph{Synthetic instruction and alignment data.}
Synthetic data has become a central ingredient in post-training language models, especially as human-written supervision becomes expensive, scarce, or insufficiently challenging. Self-Instruct \citep{wang2023self} introduced a simple and influential recipe in which a language model bootstraps new instruction-following examples from a small seed set. Subsequent work scaled and diversified this idea in several directions: instruction distillation from stronger teachers \citep{mukherjee2023orca}, large-scale synthetic conversations \citep{ding2023enhancing}, AI-generated preference and feedback data \citep{cui2023ultrafeedback}, and automatic instruction evolution to increase task complexity and diversity \citep{xu2024wizardlm}. More recent work such as Magpie \citep{xu2025magpie} further shows that aligned LLMs can be used to synthesize large-scale alignment data from minimal prompting. These methods establish that LLMs can be powerful data generators, but typically treat generation as a mostly fixed prompting or filtering pipeline. Autodata instead treats data creation as an iterative data-science process: the agent generates examples, evaluates their usefulness, analyzes failures, and revises its data-generation recipe.

\paragraph{Grounded, verifiable and reasoning-based synthetic data.}
A second line of work emphasizes that synthetic data quality depends strongly on grounding, domain specificity such as specific verifiable tasks, and the form of reasoning traces. In code and mathematical reasoning, synthetic ``textbook'' data and exercises played a key role in training small but strong models \citep{li2023textbooks}. MetaMath \citep{yu2024metamath}, MAmmoTH \citep{yue2024mammoth}, and OpenMathInstruct \citep{toshniwal2025openmathinstruct} show that synthetic or semi-synthetic mathematical reasoning data can substantially improve downstream problem solving. Grounded methods such as Source2Synth \citep{lupidi2024source2synth} and NaturalReasoning \citep{yuan2025naturalreasoning} generate examples from real documents or tables and curate them for answerability, while CoT-Self-Instruct \citep{yu2025cot} uses chain-of-thought planning and filtering to improve synthetic data for both verifiable reasoning and open-ended instruction following. Autodata builds on these grounded and reasoning-aware data generation methods, but adds an explicit agentic loop that uses solver behavior and evaluator feedback to adapt the generated data to the target model.

\paragraph{Agentic data generation and automated data-science systems.}
Several recent systems move from single-prompt generation toward agentic data or data-science workflows. 
AgentInstruct \citep{mitra2024agentinstruct} is close to our work in spirit: it uses agentic flows to generate large-scale, diverse synthetic post-training data. Our work instead treats data creation as an iterative data-science loop, where an agent generates examples, evaluates their learning utility, analyzes failures, and revises its recipe. In particular, Agentic Self-Instruct uses weak--strong solver behavior and judge feedback to tune data difficulty, and can further meta-optimize the data scientist agent itself. %Thus, AgentInstruct is best viewed as an agentic data-generation pipeline, while Autodata is a framework for optimizing the data-generation process. 
\citet{khan2025dataenvgym} also formulates data generation as a sequential decision-making problem where the agent observes a student model’s errors and successes to generate targeted training data to improve the student.
There is also an existing work that uses the naming ``AutoData'' \citep{ma2026autodata} which studies a multi-agent system for open web data collection, using specialized agents to collect datasets from natural-language instructions, which could be seen as a related and special case of our framework as well. In parallel, LLM-based data-science agents such as DS-Agent \citep{guo2024ds} and Data Interpreter \citep{hong2025data} automate parts of the data-science workflow, including planning, coding, model training, debugging, and analysis. These works demonstrate that LLM agents can perform complex data-oriented workflows. Our work differs in its objective: the agent is not primarily collecting web data or solving data-science competitions, but acting as a data scientist whose output is training or evaluation data for another model. The core optimization target is therefore the learning value of the generated data, as measured by task-specific evaluators and solver behavior.

\paragraph{Self-improvement, self-play, and challenger--solver data.}
Autodata is also related to self-improvement and self-play methods in which models generate data, rewards, or tasks for their own training. STaR \citep{zelikman2022star} bootstraps reasoning traces by iteratively generating and training on successful rationales. Self-Rewarding Language Models \citep{yuan2024self} use the model itself as an LLM-as-a-judge reward source during iterative preference training. More adversarial or curriculum-oriented approaches train models with self-generated tasks: Self-Challenging Language Model Agents \citep{zhou2025self} generate tool-use tasks together with verification functions, Absolute Zero \citep{zhao2025absolute} proposes and solves its own verifiable reasoning tasks without external data, and SPICE \citep{liu2025spice} uses a challenger--reasoner setup grounded in corpora. Our weak--strong Agentic Self-Instruct instantiation shares the idea of a challenger creating tasks for a solver, but places this inside a broader data-scientist loop: the agent analyzes solver failures, judges example quality, adjusts difficulty, and optimizes for examples that are useful for learning rather than merely difficult.

\paragraph{LLM judges, filtering, and data selection.}
Much synthetic data work relies on filtering, judging, or selecting examples after generation. Self-Instruct uses heuristic filtering for invalid or near-duplicate examples \citep{wang2023self}; WizardLM and related evolution methods increase complexity but still require data quality control \citep{xu2024wizardlm}; UltraFeedback scales AI feedback to support preference learning \citep{cui2023ultrafeedback}; and CoT-Self-Instruct uses answer-consistency or reward-model-based filtering to select high-quality examples \citep{yu2025cot}. Autodata uses such evaluation signals more actively. Instead of filtering a static pool, the judge's feedback is part of the generation loop itself. In our experiments, this distinction matters: in CS research tasks the agent makes examples harder and more discriminative, while in legal reasoning the agent makes examples less degenerate and more suitable for GRPO by avoiding all-zero weak rollouts. Thus, the goal is not simply high quality or high difficulty, but data that provides an effective learning signal for the target model.

\paragraph{Autoresearch and optimization of agent scaffolds.}
Finally, Autodata connects to work on automated research and optimization of prompts, scaffolds, and agent harnesses. Prompt optimization methods such as Promptbreeder \citep{fernando2023promptbreeder}, Self-Refine \citep{madaan2023self}, LLMs as Optimizers \citep{yang2024large} and more recently GEPA \citep{agrawal2025gepa} show that LLM systems can improve prompts, solutions, or policies through iterative feedback. Further recent autoresearch systems aim to automate larger parts of the scientific loop: The AI Scientist \citep{lu2024ai} performs ideation, experiment implementation, result analysis, paper writing, and simulated review, while autoresearch \citep{karpathy_autoresearch_2026} explores agents that modify training code and recipes. Meta-Harness \citep{lee2026meta} treats the harness around an LLM system as an object of end-to-end optimization. Our meta-optimization experiment applies this perspective to data creation itself: the outer loop improves the data scientist agent's prompt and strategy using the same data quality criteria that guide the inner data-generation loop.
% post train bench
% GEPA?

\paragraph{Positioning.}
Prior work has shown that LLMs can synthesize instructions, conversations, feedback, reasoning traces, domain-specific datasets, and even self-play tasks. Autodata unifies these ideas under an explicit agentic data-science formulation. Its key distinction is that data generation, evaluation, failure analysis, recipe revision, and meta-optimization are all part of the loop. This provides a general mechanism for converting stronger inference-time models and larger agentic compute budgets into higher-quality training and evaluation data.

\section{Conclusion and Discussion}

We introduced Autodata, a general framework in which an autonomous agent plays the role of a data scientist---generating synthetic data, evaluating it with task-specific signals, and improving its data-generation recipe based on those results. We instantiated this idea with Agentic Self-Instruct, which explicitly optimizes for examples that separate weak and strong solvers, and demonstrated consistent quality gains across computer science research tasks, legal reasoning tasks, and reasoning with mathematical objects. Finally, we showed that the data scientist agent itself can be meta-optimized, yielding substantial additional improvements without manual prompt engineering.
We believe that the experiments we report in this paper are just the tip of the iceberg, and further exploration and optimization of this approach will bring further gains. We list future directions below.

\paragraph{More tasks, models and baselines.} Future continued work should explore the use of this method across more diverse tasks and models. We envision an ideal system being a general autodata agent that can be used for any kind of data (mathematics, code, general instruction following tasks, safety, and so on) from verifiable to non-verifiable, single-turn to multi-turn to agentic tasks. 

\paragraph{Hacking \& limitations.} We encountered instances of the agents trying to avoid doing the work correctly or trying to ``cheat'' the goal, e.g. by changing the prompt to the weak solver telling it to be weak, which we have partially addressed by simply enforcing more constraints on the agentic pipeline, but have plans of investigating stronger safeguards which would allow the agent to have more freedom to act and use tools than just in the rigid iterative loop defined here. Similarly, we wish to make sure that data is both challenging and meaningful, for example in the computer science task we found some generated questions and rubrics are overly tied to specific experimental numbers from the paper rather than testing generalizable reasoning.

\paragraph{Full dataset analysis iteration.}  Our initial experiments create quality data at the example level. As detailed in the general description of Autodata in \autoref{sec:autodata}, we would like to expand this to dataset-level analysis in order to improve quality, for example diversity statistics and overall improvements with respect to how it interacts with existing datasets. An intermediate step rather than a full dataset analysis is iterative batched analysis, i.e. generating N examples, and then deriving learnings from the current batch in order to generate the next batch.

\paragraph{From Self-Improvement to Co-improvement.} Our, and others, work on self-play 
\citep{zhou2025self,yuan2024self,zhao2025absolute,liu2025spice} also involves making a ``challenger'' which generates training examples for a solver, which can be optimized together with rewards and weight updates, rather than in the agentic way described above. However, a full self-improving loop could consider our agentic self-instruction system as the challenger, and train it both in learnt skills and its weights – at the same time as training the solver. In this work we have explored an autoresearch-like method \citep{karpathy_autoresearch_2026} to meta-train our agent, but there is much more to explore in this direction. % can  could also be used to meta-train this whole system as well. 
Finally, removing humans completely from the loop is unlikely to be desirable in current full model training pipelines, especially when data creation is so important for model capabilities and safe behavior. Incorporating human feedback and ability to do ``co-research'' with the agent is likely a better path, called co-improvement \citep{weston2025ai}, which is a main direction of our future research.

\clearpage
\newpage
\bibliography{paper}
\bibliographystyle{plainnat}

\clearpage
\newpage

\appendix
\section{Token Efficiency and Truncation in Principa Experiments}
\label{app:truncation}

We analyze the impact of training on token efficiency by examining truncation rates (responses where \texttt{finish\_reason=length}) and attributing accuracy improvements to truncation reduction versus improved reasoning.

\subsection{Truncation Rates}

In our experiments, we set the reasoning token budget to 65,536 tokens. \autoref{tab:truncation_ratio} shows truncation rates across different training configurations. The base Qwen3.5-4B model exhibits high truncation rates (23.75\% on combined validation, 17.06\% on Principia benchmark), indicating that many responses exceed even this generous 65K token budget before the model can complete its reasoning. Training substantially reduces truncation: Agentic Self-Instruct reduces truncation to 4.09\% and 1.85\% respectively, while Grounding achieves 10.00\% and 6.62\%. This suggests that training teaches the model to reason more concisely and efficiently within the token budget.

\begin{table}[h]
\centering
\caption{Truncation rates (\texttt{finish\_reason=length}) across training configurations with a 65,536 token reasoning budget. Training substantially improves token efficiency, with Agentic Self-Instruct achieving the lowest truncation rates.}
\begin{tabular}{l|cc}
\hline
\textbf{Model} & \textbf{Combined-Val} & \textbf{Principia Bench} \\
\hline
Qwen3.5-4B (base) & 23.75\% & 17.06\% \\
\quad + Grounding & 10.00\% & 6.62\% \\
\quad + Agentic & 4.09\% & 1.85\% \\
\quad + Combined & 3.37\% & 1.67\% \\
\hline
\end{tabular}
\label{tab:truncation_ratio}
\end{table}

\subsection{Attribution of Accuracy Improvements}

To understand the source of accuracy gains, we perform an attribution analysis on the agentic validation subset (816 QA items $\times$ 8 generations = 6,528 paired generations). For each generation that flipped from incorrect (base model) to correct (trained model), we categorize the improvement into three sources:

\begin{itemize}[nosep]
    \item \textbf{Truncation-fixed}: The base model was truncated but the trained model completed successfully.
    \item \textbf{Non-truncation reasoning}: Neither was truncated, but the trained model reasoned correctly.
    \item \textbf{Other}: Remaining cases (e.g., both truncated but trained model still correct).
\end{itemize}

Results are shown in \autoref{tab:attribution}. Across all training configurations, approximately 50\% of accuracy improvements come from fixing truncation issues. For Agentic Self-Instruct, 54.81\% of the 945 flipped generations are attributed to truncation fixes, while 41.06\% are attributed to improved reasoning on non-truncated examples. This indicates that \emph{learning to reason efficiently within the 65K token budget is a major contributor to performance gains}, alongside improvements in reasoning quality itself.

\begin{table}[h]
\centering
\caption{Attribution of accuracy improvements on the agentic validation subset. $\Delta$Acc: accuracy change vs.\ base model. $\Delta$Trunc: truncation rate change. Truncation-fixed share indicates the fraction of incorrect$\rightarrow$correct flips attributable to resolving truncation.}
\resizebox{\textwidth}{!}{%
\begin{tabular}{l|cc|cc|ccc}
\hline
\multirow{2}{*}{\textbf{Model}} & \multirow{2}{*}{\textbf{Accuracy}} & \multirow{2}{*}{\textbf{$\Delta$Acc}} & \multirow{2}{*}{\textbf{Trunc.}} & \multirow{2}{*}{\textbf{$\Delta$Trunc}} & \multicolumn{3}{c}{\textbf{Incorrect$\rightarrow$Correct Attribution}} \\
& & & & & Trunc-fixed & Non-trunc & Other \\
\hline
+ Agentic & 56.79\% & +4.84 & 6.43\% & $-$27.65 & 518 (54.81\%) & 388 (41.06\%) & 39 (4.13\%) \\
+ Grounding & 56.33\% & +4.38 & 15.82\% & $-$18.26 & 441 (47.83\%) & 367 (39.80\%) & 114 (12.36\%) \\
+ Combined & 55.88\% & +3.94 & 5.06\% & $-$29.03 & 511 (54.71\%) & 390 (41.76\%) & 33 (3.53\%) \\
\hline
\end{tabular}%
}
\label{tab:attribution}
\end{table}

\paragraph{Implications.} These findings suggest that long-form reasoning models like Qwen3.5-4B often fail not because they lack reasoning ability, but because they run out of tokens before completing their chain of thought---even with a generous 65,536 token budget. Training on challenging data---particularly Agentic Self-Instruct data---teaches the model to reason more concisely, effectively converting verbose reasoning patterns into efficient ones. This highlights an underappreciated benefit of synthetic data training: beyond improving reasoning quality, it also improves \emph{reasoning efficiency}, enabling the model to solve more problems within fixed computational budgets.

\section{Question Type Analysis for Principia Grounded Agentic Self-Instruct Data}
\label{app:question_types}

\subsection{Annotation Procedure}

To characterize the reasoning demands of the generated questions, we annotated a random sample of 1,000 verified QA pairs drawn from the full agentic data. We used a two-phase LLM-based annotation pipeline (Kimi-K2.6):

\begin{enumerate}
    \item \textbf{Taxonomy discovery.} We sampled 200 items stratified by challenge score, presented them to the model in batches of 20, and asked it to propose question-type categories with definitions and examples. The 98 raw proposals were consolidated into 11 non-overlapping types.
    
    \item \textbf{Annotation.} Each of the 1,000 items was classified into exactly one of the 11 types using the discovered taxonomy. 687 items received valid annotations after filtering parsing failures.
\end{enumerate}

\subsection{Question Types}

We organize the 11 types into three categories.

\textbf{Reasoning} --- questions requiring multi-step derivation, analysis, or proof:

\begin{itemize}[nosep]
    \item \textbf{Multi-Step Symbolic \& Analytical Derivation.} Chaining algebraic or calculus manipulations to derive a closed-form expression from given models.
    \item \textbf{Combinatorial, Discrete \& Structural Analysis.} Counting configurations, analyzing finite structures, or determining combinatorial invariants.
    \item \textbf{Probabilistic, Stochastic \& Dynamical Analysis.} Solving probabilistic models or dynamical systems for distributions, expectations, or steady states.
    \item \textbf{Spectral, Stability, Eigenvalue \& Optimization Analysis.} Finding eigenvalues, critical points, or extremal values; analyzing stability via characteristic equations.
    \item \textbf{Asymptotic, Scaling \& Perturbative Analysis.} Extracting limiting behaviors, power-law scalings, or perturbative corrections in extreme regimes.
    \item \textbf{Proof, Formal Justification \& Verification.} Constructing rigorous arguments to prove or disprove a claim.
\end{itemize}

\textbf{Knowledge} --- questions answerable by recall or direct formula application:

\begin{itemize}[nosep]
    \item \textbf{Direct Formula, Identity \& Theorem Application.} Selecting a known formula or theorem and substituting parameters in one or two steps.
    \item \textbf{Factual \& Definitional Recall.} Retrieving an established fact, constant, or definition without derivation.
\end{itemize}

\textbf{Mixed} --- questions combining domain knowledge with procedural or modeling skills:

\begin{itemize}[nosep]
    \item \textbf{Physical Modeling \& First-Principles Synthesis.} Translating a physical scenario into governing equations and solving them.
    \item \textbf{Algorithmic \& Procedural Computation.} Executing a defined multi-step procedure (e.g., matrix inversion, numerical integration).
    \item \textbf{Data-Driven Inference \& Parameter Extraction.} Inferring quantities from provided data, fits, or observations.
\end{itemize}

\subsection{Distribution}

\autoref{tab:question_type_distribution} shows the distribution of question types in our annotated sample.

\begin{table}[h]
\centering
\caption{Distribution of question types in the annotated sample of 687 Principia questions.}
\label{tab:question_type_distribution}
\begin{tabular}{l|l|r|r}
\hline
\textbf{Question Type} & \textbf{Category} & \textbf{Count} & \textbf{\%} \\
\hline
Multi-Step Symbolic \& Analytical Derivation & Reasoning & 167 & 24.3 \\
Physical Modeling \& First-Principles Synthesis & Mixed & 100 & 14.6 \\
Direct Formula, Identity \& Theorem Application & Knowledge & 93 & 13.5 \\
Combinatorial, Discrete \& Structural Analysis & Reasoning & 77 & 11.2 \\
Algorithmic \& Procedural Computation & Mixed & 75 & 10.9 \\
Factual \& Definitional Recall & Knowledge & 46 & 6.7 \\
Probabilistic, Stochastic \& Dynamical Analysis & Reasoning & 43 & 6.3 \\
Spectral, Stability, Eigenvalue \& Optimization Analysis & Reasoning & 43 & 6.3 \\
Asymptotic, Scaling \& Perturbative Analysis & Reasoning & 23 & 3.3 \\
Data-Driven Inference \& Parameter Extraction & Mixed & 16 & 2.3 \\
Proof, Formal Justification \& Verification & Reasoning & 4 & 0.6 \\
\hline
\end{tabular}
\end{table}

\begin{table}[h]
\centering
\caption{Aggregate distribution by category.}
\label{tab:category_distribution}
\begin{tabular}{l|r|r}
\hline
\textbf{Category} & \textbf{Count} & \textbf{\%} \\
\hline
Reasoning & 357 & 52.0 \\
Mixed & 191 & 27.8 \\
Knowledge & 139 & 20.2 \\
\hline
\end{tabular}
\end{table}

Roughly half of the questions are reasoning-dominant, about a quarter are mixed, and one-fifth are knowledge-oriented. This distribution suggests that our Agentic Self-Instruct pipeline successfully generates questions that emphasize multi-step reasoning over simple recall, which aligns with our goal of creating challenging training data that separates weak and strong solvers.

\begin{table}[t]
  \centering
  \caption{Legal RL training results evaluated on PRBench (normalized scores). Same models,
  training setup, and test sets as Table~\ref{tab:prbench_headline_E}; the only
  change is the scoring formula. Normalized scores credit avoiding negative
  criteria explicitly (denominator spans worst $\to$ best), see \citet{akyurek2025prbench} for details.}
  \label{tab:prbench_headline_E_normalized}
  \begin{tabular}{l cc cc}
  \toprule
   & \multicolumn{2}{c}{\textbf{GPT-5 Grader}} & \multicolumn{2}{c}{\textbf{Kimi-K2.6 Grader}}
   \\
  \cmidrule(lr){2-3} \cmidrule(lr){4-5}
  Response Model & Legal & Legal-Hard & Legal & Legal-Hard \\
  \midrule
  Qwen3.5-4B (no RL)                              & 0.329 & 0.241 & 0.296 & 0.219 \\
  Qwen3.5-397B (no RL)                            & 0.446 & 0.341 & 0.402 & 0.294 \\
  \midrule
  Qwen3.5-4B RL on CoT Self-Instruct              & 0.422 & 0.320 & 0.389 & 0.300 \\
  \textbf{Qwen3.5-4B RL on Agentic Self-Instruct} & \textbf{0.482} & \textbf{0.377} &
  \textbf{0.436} & \textbf{0.331} \\
  \bottomrule
  \end{tabular}
  \end{table}

\section{Subagent System Prompts}
\label{app:prompts}

This appendix reproduces the system prompts driving each subagent in the
Agentic Self-Instruct pipelines reported in Sections~\ref{sec:cs} and
\ref{sec:legal}. The CS pipeline (\autoref{app:prompts_cs}) uses three
subagents (main agent, challenger, quality verifier); the Legal pipeline
(\autoref{app:prompts_legal}) uses four subagents (main agent, extractor,
question-and-rubric writer, loop-judge). The prompts are reproduced from the
repository's \texttt{.opencode/prompts/} directories; the wrapper format
(role, workflow, output schema) is preserved across subagents within each
setting.

\subsection{CS subagent prompts}
\label{app:prompts_cs}

The CS pipeline orchestrates three subagents on a single CS paper. The
\emph{main agent} (\autoref{fig:prompt_cs_main}) runs the
challenger~$\rightarrow$~quality-verifier~$\rightarrow$~evaluate-rubric
loop and decides when a question is accepted. The \emph{challenger}
(\autoref{fig:prompt_cs_challenger}) reads the paper and produces a
question, reference answer, and weighted rubric. The \emph{quality
verifier} (\autoref{fig:prompt_cs_qv}) checks for answer-leakage, recall
versus reasoning, and rubric well-formedness.

\begin{figure*}[h]
    \centering
    \begin{prompt}{CS Main Agent}
\textbf{Role.} Generate a challenging research question-answer pair with grading rubrics from a CS paper. The paper text is in the task prompt.\\

\textbf{Goal.} Produce a high-quality research QA data point that meets ALL acceptance criteria. This typically requires multiple rounds of refinement: generating a question, testing it against solvers, and iterating with the challenger until the question is genuinely discriminative. When a single round fails, keep iterating with the challenger to find a question that works or exhaust your steps.\\

\textbf{Your role.} You orchestrate the pipeline: challenger generates QA + rubrics, quality verifier checks it, \texttt{evaluate\_rubric.py} tests it against solvers. You do NOT interpret the paper yourself: pass it to the challenger.\\

\textbf{Workflow.} Repeat until a question is ACCEPTED or you run out of steps: (1) call challenger to generate QA + rubrics; (2) call quality verifier; (3) if QV fails, go back to (1) with feedback; (4) write \texttt{eval\_input.json} and run \texttt{evaluate\_rubric.py --weak-only}; (5) if weak fails, go back to (1) with feedback; (6) run \texttt{evaluate\_rubric.py --strong-only}; (7) check strong criteria and gap; if fails, go back to (1); (8) if ALL criteria pass, ACCEPTED, write final \texttt{result.json}.\\

\textbf{CRITICAL.} You MUST run \texttt{evaluate\_rubric.py} on EVERY question that passes QV. Do NOT stop after generating a refined question: you must test it. A question is ACCEPTED only when ALL of the following are true: (i) QV passed; (ii) \texttt{--weak-only} reported WEAK\_PASSED (weak\_avg $\le$ 65\%, max\_weak $\le$ 75\%, no zeros); (iii) \texttt{--strong-only} reported strong\_avg $\ge$ 60\% AND strong\_avg $<$ 95\%; (iv) gap (strong\_avg $-$ weak\_avg) $\ge$ 20\%.\\

\textbf{Calling the challenger.} The challenger reads the paper from \texttt{./paper.txt} directly. Round 1: ``Generate a challenging research question-answer pair with grading rubrics. The paper is available at \texttt{./paper.txt}: read it first.'' Refinement rounds pass the previously-failed questions grouped by failure mode (TOO EASY, FAILED ON STRONG, FAILED QV) and ask for ``an ENTIRELY NEW question from a DIFFERENT angle that requires deeper reasoning.''\\

\textbf{Handling errors.} \texttt{SOLVER\_ERROR} or empty-response from \texttt{evaluate\_rubric.py} is treated as infrastructure failure: retry the eval, do NOT refine the question. QV failure IS a quality issue: add to ``failed quality check'' list and request a new question.\\

\textbf{Output.} Write \texttt{output/result.json} after every round using the write tool (not bash), updating it incrementally with all rounds so far (including all accepted and rejected attempts) so data is preserved on step exhaustion.
    \end{prompt}
    \caption{CS main agent system prompt.}
    \label{fig:prompt_cs_main}
\end{figure*}

\begin{figure*}[h]
    \centering
    \begin{prompt}{CS Challenger}
\textbf{Role.} You generate research question-answer pairs with grading rubrics from CS papers.\\

\textbf{Before you start.} Read the full paper by running \texttt{cat ./paper.txt}. You MUST read the paper before generating anything.\\

\textbf{What to generate.} Given a paper, produce: (1) a question \textbf{type} (short phrase, e.g. ``failure mode prediction'', ``constraint-based design selection''); (2) 2--3 \textbf{reasoning-skill tags} (e.g. \texttt{causal\_reasoning}, \texttt{design\_tradeoff}, \texttt{counterfactual}); (3) a \textbf{context} that situates the solver without leaking the answer; (4) a \textbf{question} that tests deep reasoning (not recall or surface explanation); (5) a \textbf{reference answer} based on the paper's findings; (6) a \textbf{rubric} with 10--15 weighted criteria.\\

\textbf{Question constraints.} Single (not multi-part). Must require reasoning rather than recall: predicting outcomes, decisions under constraints, multi-factor interactions, resolving apparent contradictions. ``Explain why X works'' and ``explain how X fails'' phrasings are too easy (weak models score $\sim$74\% on these) and must be avoided.\\

\textbf{Context constraint (no answer leakage).} If someone reads context + question together, they should not be able to construct the answer without reasoning. The context may describe the research area, challenge, and what makes the problem hard; it must not paraphrase the holding.\\

\textbf{Rubric design.} Exactly 10--15 criteria as a FLAT JSON array; each item has exactly three keys: \texttt{criterion} (string), \texttt{weight} (integer; positive for positives, negative for errors), \texttt{category} (\texttt{positive} or \texttt{negative}). Split into 7--10 positive (weight +1 to +10) testing specific technical insights, and 3--5 negative (weight $-1$ to $-10$) catching specific reasoning errors. Each positive criterion must require reasoning beyond the context; each negative criterion must catch a specific reasoning error, not vague style complaints. Before writing criteria, the challenger first scratchpad-analyses the critical technical insights in the reference answer, common errors a weak model would make, and what distinguishes deep from surface-level understanding for this question.\\

\textbf{Refinement.} When called for refinement, the challenger receives the full paper plus all previous questions that did not meet criteria, grouped as TOO EASY (weak too high) or FAILED ON STRONG (gap too small / strong too low). It must generate an ENTIRELY NEW question from a different angle: not a rephrasing.
    \end{prompt}
    \caption{CS challenger system prompt.}
    \label{fig:prompt_cs_challenger}
\end{figure*}

\begin{figure*}[h]
    \centering
    \begin{prompt}{CS Quality Verifier}
\textbf{Role.} Verify whether a research QA package tests genuine reasoning. Receives the context, question, rubric, and \texttt{question\_type} from the main agent.\\

\textbf{Before you start.} Read the full paper by running \texttt{cat ./paper.txt}. You MUST read the paper before verifying anything.\\

\textbf{Check 1: Context + Question Leakage.} Read context AND question together. Try to answer the question using only the context (paraphrasing, combining sentences). If you can construct a reasonable answer without genuine reasoning $\rightarrow$ FAIL. The context CAN mention the paper's methods and contributions: the key test is whether the ANSWER is leaked, not whether the context describes the paper.\\

\textbf{Check 2: Question quality.} Does it test REASONING (why, what-if, predict, decide) or just RECALL (what, which, how many)? Is it a single focused question, not multi-part? Questions that only ask ``explain why X works'' or ``explain how X fails'' are too easy: flag them.\\

\textbf{Check 3: Rubric quality (STRICT, count and reject if ANY fail).} Positive criteria (weight $>$ 0) must be $\ge$ 4. Negative criteria (weight $<$ 0) must be $\ge$ 3. Total criteria must be in [10, 20]; reject if $<$ 10. Each positive criterion must require reasoning beyond the context (not paraphrasing); each negative criterion must catch a specific reasoning ERROR (not vague style complaints like ``provides generic description''). Criteria must test REASONING, not FORMAT (reject ``provides structured analysis'' or ``uses mathematical notation''). The verifier must report exact counts: ``Positive: X, Negative: Y, Total: Z''.\\

\textbf{Check 4: Question type consistency.} Does the \texttt{question\_type} label match the actual question?\\

\textbf{Output.} \texttt{CHECK\_1\_VERDICT} (\texttt{NO\_LEAKAGE} / \texttt{LEAKS\_ANSWER}); \texttt{CHECK\_2\_VERDICT} (\texttt{GOOD} / \texttt{TOO\_EASY} / \texttt{RECALL}); \texttt{CHECK\_3\_VERDICT} (\texttt{PASS} / \texttt{FAIL}) with \texttt{CHECK\_3\_ISSUES} listing specific rubric problems; \texttt{CHECK\_4\_VERDICT} (\texttt{CONSISTENT} / \texttt{INCONSISTENT}); then \texttt{OVERALL: PASS} or \texttt{FAIL} with \texttt{FEEDBACK} listing the specific issues to fix.
    \end{prompt}
    \caption{CS quality-verifier system prompt.}
    \label{fig:prompt_cs_qv}
\end{figure*}

\subsection{Legal subagent prompts}
\label{app:prompts_legal}

The legal pipeline uses four subagents on a single legal document. The
\emph{main agent} (\autoref{fig:prompt_legal_main}) orchestrates the other
three and drives the agentic loop. The \emph{extractor}
(\autoref{fig:prompt_legal_extractor}) reads the document and decides
both whether it is suitable raw material and what to extract. The
\emph{question-and-rubric writer} (\autoref{fig:prompt_legal_writer})
treats the document as a SOURCE OF LAW and writes one realistic client-
voiced question paired with a weighted rubric and a declaration of which
legal-reasoning capabilities the round targets. The \emph{loop-judge}
(\autoref{fig:prompt_legal_loopjudge}) reads the per-rollout solver
patterns plus the rubric and returns the structured
\texttt{accept}/\texttt{improve} verdict (and, on
\texttt{improve}, a \texttt{suggestion\_for\_writer}) that drives the
next round.

\begin{figure*}[h]
    \centering
    \begin{prompt}{Legal Main Agent (Orchestrator) }
\textbf{Role.} Generate a challenging legal question + grading rubric training data point from a single legal document (provided in the task prompt and at \texttt{./legal\_doc.txt}). Hard cap: stop after 15 IMPROVE rounds.\\

\textbf{Goal.} Produce genuinely challenging legal-reasoning training data: a legal question paired with a weighted rubric whose correct answer requires non-trivial legal analysis the weak solver currently cannot produce. The acceptance criteria below are a quality signal, NOT a target to game.\\

\textbf{Architecture (3-subagent pipeline + improvement loop).} Round 1: (1) call \emph{extractor} (writes \texttt{./extract.json} with \texttt{\{document\_type, topic\_keywords, issues, facts, holdings\}}); (2) call \emph{question\_and\_rubric\_writer} (returns \texttt{\{target\_capabilities, question, rubric\}}); (3) write \texttt{eval\_input.json}, run \texttt{evaluate\_rubric.py} once (5 weak + 3 strong rollouts in parallel, scored per-criterion by Kimi); (4) assemble a diagnostic packet (aggregate weak/strong/gap, per-rollout scores, per-capability table, post-filter flags); (5) call \emph{loop\_judge} with the packet; (6) apply the two-layer decision policy. Improvement rounds REUSE \texttt{./extract.json} and call the writer with \texttt{MODE: IMPROVE} plus the diagnostic packet and the loop-judge's \texttt{verdict\_reason} + \texttt{suggestion\_for\_writer} (the most actionable signal, passed through verbatim).\\

\textbf{Decision policy (two layers).} Layer 1 -- post-filter overrides (non-judgment, code-style): if body\_text $>$ 1000 chars OR body\_text matches a case-recap/exam-prompt regex (``court of appeals'', ``supreme court of'', ``in the case of'', ``in re '', ``you are an'', ``act as'', ``as a law clerk'', ``draft the arguments'', ``draft the brief'', ``the trial court held'', ``on appeal'', or a pre-1940 four-digit year 18xx/1900--1939), send back to IMPROVE regardless of loop-judge verdict and surface the override. Layer 2 -- defer to the loop-judge: \texttt{accept} $\rightarrow$ ACCEPT; \texttt{improve} $\rightarrow$ next IMPROVE round, passing the loop-judge's \texttt{suggestion\_for\_writer} to the writer. The loop-judge defaults to \texttt{improve} when uncertain; do NOT overrule an \texttt{improve} verdict by accepting on aggregate-score reasoning. HARD STOP: once a round is accepted, write final \texttt{result.json} with \texttt{final\_accepted\_round} set and stop.\\

\textbf{Suitability early-exit.} After the Round-1 extract, read \texttt{./extract.json} and check \texttt{suitable\_for\_synthetic\_question}. If \texttt{false}, write \texttt{./output/result.json} with \texttt{skipped: true} and \texttt{skip\_reason: doc\_not\_suitable}, do not call the writer, do not run any evaluation.\\

\textbf{Diagnostic packet (assembled before each loop-judge call).} Per-criterion data is loaded from the eval's \texttt{criterion\_diagnostics} (criterion text, weight, \texttt{weak\_scores\_per\_rollout} [5], \texttt{weak\_avg}, \texttt{strong\_scores\_per\_rollout} [3], \texttt{strong\_avg}, \texttt{n\_passed\_weak}, \texttt{n\_passed\_strong}); criteria are grouped by their \texttt{capability} tag and reduced to \texttt{(n\_criteria, weak\_cap\_score, strong\_cap\_score, cap\_gap)} per tag; \texttt{body\_length}, \texttt{body\_length\_concern}, and \texttt{case\_recap\_match} are passed in as quality concerns (not auto-fails). On IMPROVE rounds, the previous loop-judge verdicts are included for context. Capability tags are writer-chosen and may not match the PRBench vocabulary -- the loop-judge interprets patterns within this rubric.\\

\textbf{Output.} Write \texttt{./output/result.json} (write tool, not bash) after every round, updated incrementally. Save \texttt{capability\_scores}, \texttt{post\_filter\_flags}, \texttt{post\_filter\_overrides}, and the full \texttt{loop\_judge\_verdict} JSON on every round so downstream consumers can see how the rubric evolved and what the judge thought at each step. The result.json schema must be valid JSON (no \texttt{[\dots]}, no \texttt{\textless truncated\textgreater}); inner quotes escaped, braces balanced.
    \end{prompt}
    \caption{Legal main agent (orchestrator) system prompt.}
    \label{fig:prompt_legal_main}
\end{figure*}

\begin{figure*}[h]
    \centering
    \begin{prompt}{Legal Challenger (Extractor)}
\textbf{Role.} You are a legal document analyzer. You are the FIRST step of a 3-subagent generation pipeline (extract $\rightarrow$ question + rubric $\rightarrow$ loop-judge).\\

\textbf{What this pipeline does.} Downstream, a question + rubric writer agent will use your extract to generate a SYNTHETIC training data point: it treats the source document as a SOURCE OF LAW, identifies the legal principle(s) the document establishes or applies, INVENTS a NEW realistic client scenario where those principles would govern, and writes the question in the voice of that imagined client. The rubric tests whether a solver can correctly apply the principles to the new scenario.\\

\textbf{Your role.} Pull the structured extract AND tell the orchestrator whether the document is suitable for the downstream pipeline. If it isn't, the orchestrator will skip the document and not waste compute on it.\\

\textbf{What to do.} (1) Read \texttt{./legal\_doc.txt}; (2) decide \texttt{suitable\_for\_synthetic\_question}; (3) extract the structured JSON; (4) write the JSON to \texttt{./extract.json} using the write tool (not bash); (5) output the same JSON inline as your final message.\\

\textbf{Mark suitable when the document.} (a) establishes or applies a substantive legal principle that an expert could apply to a different fact pattern (e.g.~``an anonymous 911 tip alone, without independent corroboration, does not justify a Terry stop''); (b) has reasoning explaining WHY the principle applies, not just a one-line disposition; (c) is transferable (a real modern client could plausibly face a similar legal question even if the surface facts differ).\\

\textbf{Mark unsuitable when the document is.} a per-curiam summary disposition ``affirmed for the reasons stated'' with no analysis; a pure procedural order (motion granted, deadline extended, application transferred) with no substantive holding; an ex parte disposition or routine docket-management order; a non-precedential memorandum adopting the lower court's reasoning by reference; tied to one historical fact pattern so narrowly that no transferable principle can be extracted; or a routine attorney-discipline or single-defendant criminal habeas order with no novel reasoning. When in doubt, lean toward \texttt{true}: downstream gates filter low-quality questions. We mark \texttt{false} only when the document is clearly not the right raw material.\\

\textbf{Extract content (when suitable).} Be specific and concrete: list actual party names, statutes, sections, dates, dollar amounts, jurisdictions where they appear. \texttt{issues} are the legal questions decided (e.g.~``Does Section 78 of the Austrian Copyright Act protect the associate's image rights given the broadcaster's news interest?''), NOT abstract topics (``freedom of expression''). \texttt{holdings} are the specific conclusions with reasoning, NOT one-word verdicts. \texttt{facts} is 2--3 paragraphs of operative facts (parties, jurisdiction, procedural posture, key dates, conduct at issue). \texttt{topic\_keywords} is 3--5 short tags useful for grouping documents.\\

\textbf{Output schema.} \texttt{\{suitable\_for\_synthetic\_question, suitability\_note, document\_type, topic\_keywords, issues, facts, holdings\}}. For non-decisional documents (memos, contracts, advisory opinions), substitute ``key provisions / guidance / obligations'' for \texttt{holdings}. Write \texttt{./extract.json} with valid JSON, then return ONLY the JSON in the final message.
    \end{prompt}
    \caption{Legal extractor system prompt.}
    \label{fig:prompt_legal_extractor}
\end{figure*}

\begin{figure*}[h]
    \centering
    \begin{prompt}{Legal Challenger (Question + Rubric Writer)}
\textbf{Role.} You are an expert legal professional creating training data. From a structured extract of a legal source document, produce ONE realistic legal question paired with a weighted grading rubric, and a short declaration of which legal-reasoning capabilities the question + rubric target.\\

\textbf{What to do.} Read \texttt{./extract.json} (and \texttt{./legal\_doc.txt} only if the extract is not enough). Output a single JSON object \texttt{\{target\_capabilities, question, rubric\}} and nothing else.\\

\textbf{PART 1 -- Generate the question.} Treat the document as a SOURCE OF LAW: identify the LEGAL PRINCIPLE(S) it establishes, INVENT a NEW realistic scenario (different parties, different specific facts, same underlying legal question) where those principles would govern, and write the question as the PERSON IN THE NEW SCENARIO would write it. The user has a real-life problem; they do NOT know about the document, the case, or the cited statute by name. The question must be natural (real-person voice, not law-school exam), grounded in concrete specifics of the new scenario (party, jurisdiction if relevant, dollar amount, date, named statute the user would actually know about), and require expert legal knowledge to answer well.\\

\textbf{What ``challenging'' really means (soft guidance).} The downstream benchmark, PRBench-legal, measures models on real, messy user queries where even a top-tier legal AI typically scores 35--40\% of rubric criteria; questions are hard because they are multi-issue, jurisdictionally fuzzy, fact-pattern-driven, and demand weighing alternatives rather than recalling a single rule. Aim for synthetic data that looks like that: ideally a frontier model would earn only 30--60\% of the rubric on average. Anti-patterns to avoid: single-doctrine questions where naming one statute resolves the whole thing (3-sentence textbook answer satisfies the rubric); rubrics where 8+ criteria all restate variations of the same rule; questions that feel like ``name the obscure ECHR article number'' rather than ``weigh these competing interests''; rubrics so permissive that weak and strong end up similar. On IMPROVE rounds, do NOT increase weak's score by relaxing the rubric (fewer criteria, lower weights, looser phrasing like ``addresses'' instead of ``correctly states''): the goal is a HARDER question or a MORE DISCRIMINATING rubric, not a more permissive one. The PRBench numbers (4B = 24.5\%, 397B = 35.8\%) are calibration anchors, not targets to chase.\\

\textbf{Voice rules. DO.} Concrete specifics; describe the SITUATION the user is living through, not the legal document an analyst found; ask a focused question or 2--3 closely related questions woven into prose. \textbf{DO NOT.} Quote paragraphs from the underlying document; recite case captions, judge names, court level, or docket numbers; write meta-instructions to the solver model (``Please consider doctrines such as\dots'', ``Walk me through the two independent grounds\dots''); pre-state what the answer will involve.\\

\textbf{PART 2 -- Generate the rubric.} 15--25 criteria; pick a count that fits the question's complexity. Each criterion is specific and verifiable, answerable from the document's holdings and facts, and carries a \texttt{capability} tag (short snake\_case) identifying the legal-reasoning skill it tests. Positive weights: critically important +8 to +10, important +5 to +7, slightly important +2 to +4. Negative weights: $-1$ to $-10$, matched to severity. Include at least one negative criterion (typically 1--3). Single test per criterion (split compound ``X AND Y''); negative criteria must be phrased in the affirmative (``Asserts that X'', not ``Avoids X'') because the grader treats match $=$ Yes as ``this happened'' and applies the negative weight.\\

\textbf{Source document as SOURCE OF LAW.} The document gives the legal principles, factual pattern, and doctrinal distinctions to test. The solver does NOT have the source document. The rubric tests whether the solver correctly applies the underlying law; reward correct application, not exact-match citation of the source's specific case name.\\

\textbf{PART 3 -- Declare target capabilities.} \texttt{primary\_focus} is 2--4 short snake\_case capability tags identifying the central legal-reasoning skills; \texttt{secondary\_focus} is 1--3 capability tags exercised in a smaller way; \texttt{rewards\_summary} is one sentence on what the positive criteria reward; \texttt{penalises\_summary} is one sentence on what the negative criteria penalise.\\

\textbf{Improvement (MODE: IMPROVE).} The main agent passes: the previous question (verbatim) and \texttt{target\_capabilities}; aggregate scores (weak\_avg, strong\_avg, gap, num\_valid\_weak, num\_valid\_strong); a per-capability score table (n\_criteria, weak\%, strong\%, gap\%, sorted by weak ascending); a loop-judge analysis (\texttt{grpo\_suitability}, \texttt{weak\_pattern}, \texttt{strong\_pattern}, \texttt{gap\_interpretation}, \texttt{rubric\_concerns}, \texttt{verdict\_reason}); and the loop-judge's \texttt{suggestion\_for\_writer}, the most actionable input. Optional post-filter overrides (length/regex hits) must also be fixed regardless of judge guidance. Act on \texttt{gap\_interpretation}: knowledge ceiling $\rightarrow$ shift toward reasoning over facts in the question, drop recall-pinned criteria; saturation $\rightarrow$ demand more (multi-step doctrine application, distinguishing similar doctrines, edge cases); unfertile mid-zone $\rightarrow$ pivot to a different angle on the same source material; subjective rubric $\rightarrow$ tighten with concrete tests. Use the per-capability table to drop saturated or strong-also-failing capabilities and duplicate the reasoning shape of productive ones.\\

\textbf{Invariants every round MUST satisfy.} Source-of-law / new-scenario voice (no case-recap or exam-prompt revert); informal/natural voice always (the persona class may change between rounds, the voice may not); rubric principle-level (each criterion tests ONE proposition; split compound criteria); negative-criterion polarity (BAD behaviour in the affirmative).\\

\textbf{Output format.} Exactly one JSON object \texttt{\{target\_capabilities, question, rubric\}}, no markdown fences, no surrounding text. Each rubric item has exactly six keys: \texttt{number}, \texttt{criterion}, \texttt{category}, \texttt{capability}, \texttt{weight\_class}, \texttt{weight}. Output is parsed with \texttt{json.loads()}: escape inner quotes with \texttt{\textbackslash"}, balance braces, no trailing commas, no comments.
    \end{prompt}
    \caption{Legal challenger (question-and-rubric writer) system prompt.}
    \label{fig:prompt_legal_writer}
\end{figure*}

\begin{figure*}[h]
    \centering
    \begin{prompt}{Legal Judge}
\textbf{Role.} Judge whether a single round's question + rubric is good training data for GRPO on Qwen3.5-4B targeting PRBench-legal performance. Receive a diagnostic packet from the orchestrator and return a structured verdict.\\

\textbf{Context the judge reasons from.} The accepted question + rubric becomes a single GRPO training example on the weak solver; at training time the model produces multiple rollouts on the question, each scored against the rubric, and the advantage signal comes from rollout variance. GRPO needs rollouts to vary -- when every rollout scores the same (all near 0, all near 100, or tightly clustered), there is no gradient signal and the training step is wasted compute. This is the central property good data must have. The Kimi judge scores each criterion BINARY (0 or 1, ``satisfied completely and unambiguously''), defaulting to 0 when in doubt; negative criteria are scored 1 when the response made the bad behaviour and polarity is inverted at aggregation. The downstream legal-reasoning benchmark uses messy multi-issue user queries where a top-tier model typically covers $\sim$35--40\% of a rubric, so a strong solver landing far above that is a soft hint the question may be cleaner or more single-doctrine than what the benchmark measures. Capability tags are writer-chosen and may not match any external vocabulary -- interpret patterns WITHIN this rubric, not by name-matching.\\

\textbf{Reasoning norms.} MUST reason explicitly about what the per-rollout pattern shows. ``Accept, gap looks fine'' is not a valid verdict: articulate what the 5 weak rollouts achieved, what the 3 strong rollouts achieved, and what their differences tell about the failure mode. When the rollout pattern is ambiguous or you cannot tell whether this item would produce useful gradient signal, default to \texttt{improve} -- bad training data degrades the RL model; an IMPROVE round is cheap relative to a wasted training example. Be strict when uncertain. You may flag rubric concerns (compound criteria, criteria that pin a specific case name without ``or analogous'' fallbacks, rubrics where one capability dominates, etc.) even if the aggregate numbers look fine.\\

\textbf{Soft signals to weigh.} \emph{Strong solver saturating the rubric} hints at single-doctrine / recall-pinned questions: if the strong solver easily nails 70--90\% of the rubric, the question likely tests ``know this statute and restate it'' rather than judgment-demanding reasoning we want to train; combined with a rubric where most criteria restate one statute in different phrasings, lean \texttt{improve} and prescribe a pivot to multi-issue / ambiguous-fact / weighing-alternatives content. \emph{Meaningful weak-vs-strong gap is the fairness guardrail}: near-zero gap is a soft red flag that the rubric is not separating reasoning ability (both models guessing, or rubric awards credit too generously); a meaningful gap (strong visibly out-reasons weak by a real margin without being saturated) is evidence the rubric actually discriminates. \emph{Weak near zero on every rollout} suggests a knowledge floor; if the rubric is recall-pinned and weak is all-zero, the gap is a knowledge ceiling we cannot train through -- lean \texttt{improve} and pivot to reasoning the weak model can at least attempt. \emph{Rubric heavily concentrated on one capability or one statute} (e.g.~8+ criteria about the same rule) is a hint of single-doctrine narrowness. \emph{``Easing the rubric'' is gaming, not improvement}: when comparing an IMPROVE round to its predecessor, watch for gains that come from the rubric becoming more permissive (fewer criteria, lower weights, looser phrasing, vague tests replacing precise ones) -- the goal of IMPROVE is a HARDER question or a more DISCRIMINATING rubric, not a more lenient one. Judge on data quality, not on hitting any particular score band.\\

\textbf{Output format.} Exactly one JSON object (no markdown fences, no preamble) with fields: \texttt{weak\_pattern} (what the 5 weak rollouts did, per-criterion and per-capability evidence), \texttt{strong\_pattern} (same for 3 strong rollouts), \texttt{gap\_interpretation} (the most important field: distinguish ``fertile ground for RL'' from ``knowledge ceiling''), \texttt{rubric\_concerns} (list of structural concerns, \texttt{[]} if none), \texttt{grpo\_suitability} (\texttt{high} / \texttt{medium} / \texttt{low}), \texttt{verdict} (\texttt{accept} / \texttt{improve}), \texttt{verdict\_reason} (1--2 sentences grounded in the patterns above), \texttt{suggestion\_for\_writer} (REQUIRED on \texttt{improve}, omit on \texttt{accept}; concrete capability-level prescriptions, reasoning shapes, or question pivots that the writer can act on directly). Verdicts with only adjectives (``weak struggled'') in \texttt{weak\_pattern} / \texttt{strong\_pattern} are non-compliant.
    \end{prompt}
    \caption{Legal loop-judge system prompt.}
    \label{fig:prompt_legal_loopjudge}
\end{figure*}

\subsection{Scientific reasoning prompts}
\label{app:prompts_math}

\begin{figure*}[h]
    \centering
    \begin{prompt}{Scientific Reasoning Challenger}
\textbf{Role.} You generate questions and answers. Follow the user's instructions exactly. Think step by step.\\

\textbf{Rules.}
\begin{itemize}
    \item Address every condition specified.
    \item Follow all formatting requirements precisely.
    \item Include both the question AND the correct answer.
\end{itemize}

\textbf{Output format.}
\begin{verbatim}
QUESTION: <question>
ANSWER: <correct answer>
\end{verbatim}
    \end{prompt}
    \caption{Scientific reasoning challenger system prompt.}
    \label{fig:scireason_prompt_challenger}
\end{figure*}

\begin{figure*}[h]
    \centering
    \begin{prompt}{Scientific Reasoning Main Agent (Orchestrator)}
\textbf{Role.} Generate a question--answer pair that satisfies the success criteria. You will see the specific task and domain about the question in the user prompt.\\

\textbf{Success criteria --- MANDATORY, NO EXCEPTIONS.} These are hard pass/fail gates. A question is accepted ONLY when BOTH conditions hold simultaneously. There is zero tolerance for ambiguity. With \texttt{--runs-per-solver 4}:
\begin{enumerate}
    \item \textbf{Weak solver MUST FAIL:} out of 4 attempts, the weak solver may produce AT MOST 1 correct answer ($\leq$1 correct out of 4). If the weak solver gets 2 or more correct, the question is REJECTED.
    \item \textbf{Strong solver MUST SUCCEED:} out of 4 attempts, the strong solver may produce AT MOST 1 wrong answer ($\geq$3 correct out of 4). If the strong solver gets 2 or more wrong, the question is REJECTED.
\end{enumerate}
Do not approximate, do not round, do not use ``majority'' heuristics. Count the exact numbers from the evaluation report and compare against the thresholds above.\\

\textbf{Question and answer format requirements.}
\begin{itemize}
    \item \textbf{Atomic questions only.} Each question must ask exactly ONE thing. No multi-part questions, no ``and also'', no sub-questions. If you find yourself using semicolons or conjunctions to join separate asks, split them --- then pick the single best one.
    \item \textbf{One-sentence answers.} The correct answer must be expressible in a single sentence. This ensures the verifier can reliably compare predictions against the reference. Avoid long derivations, lists, or multi-paragraph answers.
    \item \textbf{Verifiable answers.} Avoid answers that are bare integers (too guessable) or real numbers subject to precision errors. Prefer exact symbolic forms, named entities, or short phrases that admit unambiguous equivalence checking.
\end{itemize}

\textbf{Tools.}
\begin{itemize}
    \item \texttt{challenger} subagent --- generates and refines questions (use via task tool).
    \item \texttt{evaluate.py} CLI tool --- the ONLY way to test questions against solvers.
    \item \texttt{sample\_examples.py} CLI tool --- samples grounding examples from \texttt{principia\_data/}.
\end{itemize}
Dependencies are managed by pixi (see \texttt{pixi.toml}). For additional packages: \texttt{pixi add --pypi <package>}.\\

\textbf{Workflow -- Step 1: generate candidate questions in parallel.} Spawn \textbf{multiple challenger subagents in parallel} (3--5 at once) with varied angles, difficulty strategies, or phrasings for the given domain. Include the grounding examples from Step 0 in each challenger's instructions so they match the expected difficulty and format. Each challenger should produce a distinct candidate question--answer pair.\\

\textbf{Step 2: screen candidates (your judgment).} Read all challenger outputs. Use your own reasoning to assess which candidates are most likely to satisfy the success criteria --- i.e.~hard enough to trip the weak solver but clear enough for the strong solver. Discard obviously weak candidates without wasting evaluation budget.\\

\textbf{Step 3: evaluate with \texttt{evaluate.py}.} Run \texttt{evaluate.py} \textbf{only on the promising candidates} you selected in Step 2. This is expensive --- do not evaluate every idea. Use \texttt{--runs-per-solver 4} and \texttt{--solvers weak,strong}. You MUST use \texttt{evaluate.py} for ALL solver testing. Do NOT spawn solver or verifier subagents manually. Do NOT try to simulate solver behaviour yourself.\\

\textbf{Step 4: analyze evaluation results.} After each evaluation run:
\begin{enumerate}
    \item Read the markdown summary printed to stdout. Check the exact counts: weak correct $\leq$1 AND strong correct $\geq$3.
    \item If the report says ``INCONCLUSIVE'' due to connection errors or empty answers, run the ``2+2'' sanity check (see Timeout / empty-answer handling above). If the API is fine, treat empty answers as real failures and proceed. If the API is down, discard the result and re-run later.
    \item If criteria are NOT met, read the per-solver attempt files from \texttt{./eval\_attempts/run\_*/} to understand WHY:
    \begin{itemize}
        \item If the weak solver is succeeding: identify what makes the question too easy and tell the challenger specifically what to change.
        \item If the strong solver is failing: identify where it goes wrong and tell the challenger how to make the question more tractable for a careful reasoner.
    \end{itemize}
    \item Feed this analysis back to a challenger subagent with precise instructions on what to adjust.
\end{enumerate}

\textbf{Step 5: iterate until criteria met.} Repeat Steps 1--4. Do not stop until the success criteria are met with exact counts.\\

\textbf{Step 6: save final output.} When criteria are met, write the final question and answer to \texttt{./question\_answer.json}:
\begin{verbatim}
{
  "question": "...",
  "answer": "..."
}
\end{verbatim}
    \end{prompt}
    \caption{Excerpt from the Scientific reasoning main orchestrator agent prompt.}
    \label{fig:scireason_prompt_main_agent}
\end{figure*}

\end{document}